\documentclass{article}

\usepackage[final]{corl_2026} 

\usepackage[utf8]{inputenc} 
\usepackage[T1]{fontenc}    
\usepackage{inconsolata}

\usepackage{algpseudocode}  
\usepackage[ruled, linesnumbered, boxed]{algorithm2e}
\usepackage{parskip}
\usepackage{changepage}

\usepackage{color}
\usepackage[colorlinks,
            linkcolor=blue,
            anchorcolor=blue,
            citecolor=green,
            urlcolor=Aqua,
            backref=page]{hyperref}
\usepackage{wrapfig}
\usepackage{caption}
\usepackage{subcaption}
\usepackage{cancel}
\usepackage{graphicx}
\usepackage{epsfig} 
\usepackage{times} 
\usepackage{booktabs}  
\usepackage{amsmath}
\usepackage{amssymb}  
\usepackage{dsfont}
\usepackage[table, dvipsnames, svgnames, x11names]{xcolor} 
\usepackage{marvosym}
\definecolor{mygray}{gray}{.9}
\usepackage{pifont}
\usepackage{nicefrac}       
\usepackage{microtype}      
\usepackage[table, dvipsnames, svgnames, x11names]{xcolor}        
\usepackage{colortbl}

\usepackage{multirow}
\usepackage{multicol}
\usepackage{cite}

\usepackage{enumitem}
\usepackage{graphics} 
\usepackage{svg}
\usepackage{bm}
\usepackage{float} 
\usepackage{stfloats}
\usepackage{adjustbox}
\usepackage{tablefootnote}
\usepackage[para]{footmisc}

\title{HiRE: Hindsight Reward Editing for Policy Finetuning}

\author{
Haoyi Niu$^{1}$\thanks{Equal contribution, order determined by coin flip and may be listed in either order. Correspondence to: \texttt{niu@berkeley.edu}.}\  , Zhengtao Han$^{1*}$, Yufeng Ji$^{2}$, Zhongyu Li$^{2}$, Koushil Sreenath$^{1}$\\
$^1$ University of California, Berkeley,
$^2$ The Chinese University of Hong Kong\\
}

\begin{document}

\maketitle
\thispagestyle{empty}

\begin{abstract}
Pre-trained robot policies always require finetuning to adapt to specific environments. 
Reinforcement Learning (RL) offers high performance potential because it improves action optimality rather than simply mimicking data. 
However, such potential depends heavily on reward quality.
Sparse rewards lack process feedback, human-designed rewards are costly and biased, and semantic rewards from foundation representations are often not control-centric.
We propose Hindsight Reward Editing (HiRE), a training-free framework to break this reward bottleneck. 
HiRE bridges the broad knowledge of foundation representation models with physical control awareness, by contrasting successful and failed trajectories in hindsight. 
It calibrates foundation representation models by identifying ``trap states'' that are predicted as high-rewarding states yet eventually result in failure, and vice versa.
HiRE explicitly penalizes these traps while boosting rewards for critical successful states. This approach can be flexibly compatible with any foundation representations and RL algorithms.
Experiments show that HiRE consistently outperforms other reward recipes by delivering dense, control-aware feedback that prevents value function collapse and reward hacking, thereby achieving superior sample efficiency, stable policy updates, and higher performance ceilings, e.g., at least $3\times$ performance of the base policies.
Qualitative results at \href{https://hire-project.github.io/}{Project Website}.

\end{abstract}

\keywords{Reinforcement Learning, Reward Specification, Policy Finetuning}

\section{Introduction}
Reinforcement Learning (RL) has drawn renewed interest in post-training robot policies~\citep{intelligence2025pi,xu2026rl}.
By optimizing for long-term value, RL policies are naturally robust to perturbations and demonstrate strong generalization during deployment. 
This capability stems from reward signals, which allow RL to ground policies in task objectives rather than merely mimicking expert demonstrations.
However, reward specification remains a long-standing bottleneck of robot learning.
Sparse rewards fall short on long-horizon tasks because they lack detailed process feedback.
To address this, many works leverage human priors to specify dense rewards, which inevitably introduce subjective bias and struggle to generalize across diverse tasks.
Unsatisfied with using only task-specific rewards, emergent efforts focus on training expressive reward models that offer fine-grained signals for unseen tasks.
Yet, obtaining high-quality reward labels for grounding such large-scale models remains a significant bottleneck.
Therefore, current general-purpose reward models still struggle with out-of-distribution (OOD) tasks~\citep{lee2026roboreward,liang2026robometer,chen2026topreward} and often require further calibration to faithfully adapt to new scenarios~\citep{tan2025robo}.

These limitations point to two key desiderata for robot reward modeling: generalization across diverse visual scenarios and adaptation to task-specific physical outcomes. For generalization, foundation vision(-language) representations can be directly reused to produce similarity-based rewards in a pretrained latent space~\citep{rocamonde2024vision,ma2022vip,ma2023liv}. These models are trained on massive amounts of image(-text) data and thus they seldom suffer from OOD issues, while they are not control-aware because they were never trained on robot behavioral data~\citep{oquab2024dinov,tschannen2025siglip}. For adaptation, reward signals should be grounded in physical task outcomes during policy finetuning, using interactive feedback without expensive human labels or additional reward-model training. Thus, a natural question arises: \textit{Can we build a reward scheme that is both universal to generalize and cheap to adapt to new scenarios?} A promising solution would enjoy the broad generalization of foundation vision(-language) representations with minimalist efforts of grounding rewards to task-specific physical outcomes during policy finetuning.

Instantiating such a solution requires addressing two major challenges.
Primarily, similarity-based rewards based on foundation representation models often misvalue robot states~\citep{biza2025robot}.
For example, it might highly reward a robot for moving toward a placement area even if it has already dropped the object that it should take to that place. This happens because the model prioritizes visually dominant robot motions over small but critical objects, failing to capture the true task objectives.
Besides, na\"ively adjusting rewards during policy finetuning can destabilize training.
A constantly shifting reward landscape often destroys the ability of RL policies to reason about long-term value that helps navigate to final task completion~\citep{rolnick2019experience}.
To address these, we instantiate \textbf{Hi}ndsight \textbf{R}eward \textbf{E}diting (\textbf{HiRE}), a training-free reward recipe that adjusts rewards produced by foundation vision(-language) representations by contrasting successful and failed trajectories in hindsight of interactions during policy finetuning.
HiRE derives a closed-form formulation that directly calibrates reward landscape: it penalizes states that are too close to failure traps and boosts states that lead to success.
This approach ensures the rewards are grounded in actual physical outcomes without requiring any extra gradient-based optimization.
Furthermore, we incorporate these adjusted rewards through potential-based reward shaping~\citep{ng1999policy}. 
By combining sparse rewards that indicate success with our refined dense rewards, we provide control-aware feedback while mathematically guaranteeing that optimal policy for task success remains unchanged. 
These designs allow HiRE to break the control-unawareness bottleneck of foundation representation models with minimal computation and high efficiency.

In summary, our contributions are three-fold:
(1) \textbf{A practical and scalable reward recipe for real-world robot policies.}
We bridge the gap between the strong generalization of foundation representations and the specific task requirements of physical control.
By using interaction data from specific tasks, HiRE provides the ``missing piece'' of control awareness through a training-free reward editing paradigm that empowers sample-efficient RL.
It can be highly compatible with any foundation representation models like DINOv2~\citep{oquab2024dinov} and SigLIP~\citep{tschannen2025siglip}, as well as any RL finetuning algorithmic designs summarized in DICE-RL~\citep{sun2026prior}.
(2) \textbf{Strong empirical results on simulation and real-world tasks.}
HiRE achieves strong performance and high sample efficiency on prehensile, non-prehensile, and long-horizon manipulation tasks on MimicGen~\citep{mandlekar2023mimicgen} and RoboMimic~\citep{mandlekar2022matters} where it can improve low base success rates ($<10\%$) to near 100\%, as well as on a single YAM robot arm where it achieves at least $3\times$ improvement sometimes with only 30 online rollouts.
(3) \textbf{In-depth analyses on effects of HiRE. }
We provide systematic studies to understand why HiRE works, including visualizing the evolving reward landscapes during policy finetuning, and ablation studies on the key components.






\section{Related Work}
\subsection{Reward Learning for Robotics}\label{sec:reward-modeling}

Learning effective reward models remains a long-standing bottleneck in RL for robotics. One line of work learns task-specific reward models from explicit supervision. Methods such as DrS~\citep{mu2024drs} and SARM~\citep{chen2026sarm} require intensive human annotations of stages or task progress, while Rank2Reward~\citep{yang2024rank2reward} leverages rankings from passive videos to learn shaped reward functions. Recent methods further improve trajectory-level reward evaluation: RoboReward~\citep{lee2026roboreward} and RoboMeter~\citep{liang2026robometer} introduce dedicated metrics for trajectory comparison and progress tracking, while Robo-Dopamine~\citep{tan2025robo} learns a reward model from large-scale multi-view manipulation data. Despite their effectiveness, these methods often require task-specific supervision, reward-model training, or large-scale curated robotic data, limiting their generalization to novel visual scenes and task semantics. A complementary line of work derives zero-shot rewards from embedding-space similarity between current and goal observations. Representations such as VIP~\citep{ma2022vip}, LIV~\citep{ma2023liv}, R3M~\citep{nair2023r3m}, and DecisionNCE~\citep{li2024decisionnce} are pretrained on passive behavior datasets so that representation distance reflects task progress. However, since these models can only learn from passive expert data, they often suffer from OOD generalization issues in unseen scenarios. To address this, GCR~\citep{biza2025robot} extends VIP with a goal-contrastive objective trained on task-specific successful and failed trajectories to finetune embeddings during deployment.

To fundamentally bypass such OOD vulnerabilities, emergent efforts repurpose large-scale foundation vision(-language) representations trained on internet-scale, non-robotic data to achieve robust open-world generalization. 
For instance, CLIPReward~\citep{rocamonde2024vision} extracts language-conditioned rewards from CLIP~\citep{radford2021learning} by evaluating the cross-modal semantic alignment between text prompts and robot visual observations. 
RewardingDINO~\citep{krack2026rewarding} injects task-specific semantics by training an explicit preference-based reward head on top of frozen DINOv2 embeddings~\citep{oquab2024dinov}. 
TOPReward~\citep{chen2026topreward} instead extracts training-free rewards from from pretrained vision-language models (VLMs) by scoring task-completion token probabilities given trajectory prefixes and instructions.
Although these approaches enjoy exceptional generalization ability of foundation models, they are inherently \emph{control-unaware} due to the lack of physical interaction data in their pretraining phase. They evaluate task progress strictly through passive visual matching, making them highly prone to overestimating deceptive states where visually similar scenes correspond to opposite physical outcomes.
HiRE achieves training-free reward editing while remaining control-aware. It calibrates foundation representation models in hindsight by contrasting successful and failed rollouts, which introduces no auxiliary network and no manual dense labels, and is agnostic to choices of representation models and RL algorithms.


\subsection{RL Policy Finetuning for Robotics}
Recent work has increasingly used RL in post-training stage for pretrained robot policies~\citep{intelligence2025pi_}. One line tailors specialized RL recipes to the structure of generative policies: $\pi_{\texttt{RL}}$~\citep{chen2025pi_} makes flow policies amenable for direct finetuning by PPO~\citep{schulman2017proximal} via constructing tractable likelihoods. 
Direct finetuning is often expensive for fresh samples, so DSRL~\citep{wagenmaker2025steering} instead keeps the base policy frozen and only steers in its latent-noise space.
Moreover, ResFiT~\citep{ankile2025residual} and Policy Decorator~\citep{yuan2025policy} attach residual modules that learn additive offsets to the base policy output, whereas RL-Token~\citep{xu2026rl} distinguishes itself from traditional residual methods by training a lightweight policy head for final actions upon a compact token readout of a Vision-Language-Action (VLA) model.
DICE-RL~\citep{sun2026prior} instead freezes the base policy and treats it as a stochastic action proposal, contracting its distribution toward high-value modes through a residual actor~\citep{ankile2025residual,yuan2025policy} and value-guided action selection~\citep{nakamoto2025steering}. 
While all these methods focus on policy improvement, HiRE is entirely orthogonal as it addresses the reward improvement. Consequently, HiRE can be seamlessly integrated with any of these policy finetuning methods, lifting their performance ceilings by providing dynamic, training-free reward shaping that guides policy optimization toward physically correct trajectories.

\section{Preliminaries}
We formulate the RL problem as a finite-horizon Markov Decision Process~(MDP)~\citep{sutton1998introduction}, defined by a tuple $\mathcal{M}:=\left(\mathcal{S}, \mathcal{A}, R, \mathcal{P}, \gamma, H\right)$. $\mathcal{S}$ and $\mathcal{A}$ denote the state and action space, $R: \mathcal{S}\times\mathcal{A}\times\mathcal{S}\rightarrow\mathbb{R}$ represents the reward function, $\mathcal{P}:\mathcal{S}\times\mathcal{A}\rightarrow\Delta(\mathcal{S})$ stands for the transition dynamics, $\gamma\in[0,1]$ is the discount factor and $H$ is the task horizon. 
The goal of RL is to find the optimal policy $\pi^*$ that maximizes cumulative discounted reward starting from an initial state distribution $\rho$, \emph{i.e.}, 
$\pi^* =\arg\max_{\pi} \mathbb{E}_{s_0\in\rho,a\sim\pi(\cdot|s), s^\prime\sim \mathcal{P}(\cdot|s, a)}\left[\sum_{t=0}^{H} \gamma^{t} R\left(s, a\right)\right]$. RL methods based on approximated dynamic programming typically learn an action-value function $Q(s,a)$, and optionally, a state value function $V(s)$ to practically estimate the cumulative discounted reward for policy optimization.


\section{HiRE: A Training-free Reward Recipe for Policy Finetuning}
Our objective is to build a control-aware reward function from semantic foundation representations without any additional training. To achieve this, we leverage success and failure signals in past interactions to dynamically reshape the reward landscape in hindsight. 

Ideally, an optimal control-aware reward should guide the robot away from failure traps and toward successful paths. In adversarial imitation learning~\citep{ho2016generative,fu2018learning}, a standard binary cross-entropy objective is widely used to distinguish expert demonstrations from policy rollouts. Drawing inspiration from this, we can analogously formulate a discriminator $D(s)$ to distinguish the success manifold from the failure one. This discriminator is trained to predict the probability that a given state $s$ belongs to a successful trajectory: $\max_{D} \mathbb{E}_{s \sim p_{+}}[\log D(s)] + \mathbb{E}_{s \sim p_{-}}[\log(1 - D(s))]$
where $p_{+}(s)$ and $p_{-}(s)$ denote the state distributions of successful and failed states, respectively. According to Maximum Entropy Inverse RL~\citep{ziebart2008maximum}, the optimal reward function $R^*(s)$ is mathematically tied to the log-odds of this perfect discriminator $D^*(s)$:
\begin{equation}
R^*(s) = \log \left( \frac{D^*(s)}{1 - D^*(s)} \right) = \log p_{+}(s) - \log p_{-}(s)\label{eq:irl}
\end{equation}
This formulation delivers a key insight: \textbf{the ideal reward is simply the log-density ratio between success and failure.} 
Thus, rather than training a parameterized discriminator through costly optimization, we can analytically determine the optimal reward if these densities can be evaluated directly. 
This shifts the core challenge from neural network training to density estimation, opening the door for a training-free, non-parametric solution within the representation space.

\subsection{Non-parametric Reward Estimation in Foundation Representation Space}\label{hire}
To completely circumvent expensive model training, we estimate the densities $p_+(s)$ and $p_-(s)$ using non-parametric Kernel Density Estimation (KDE). Under the general KDE framework, the probability density at any state $s$ is evaluated by averaging the kernel similarities to all historical samples within a support set: $p(s) \approx \frac{1}{N} \sum_{i=1}^N K(s, s_i)$.
A major challenge is that frozen vision foundation models produce representations that reside on a high-dimensional unit hypersphere due to $L_2$ normalization. Traditional Euclidean kernels perform poorly in such spaces because they ignore the curved geometry of the manifold. To respect this geometric property, we employ the Von Mises-Fisher (vMF) kernel~\citep{mardia2009directional}, which is specifically designed for directional data on hyperspheres:
\begin{equation}
K(s, s_i) = C(\kappa) \exp\left( \kappa \cdot \text{sim}_\phi(s, s_i) \right)
\end{equation}
where $\text{sim}_\phi(s, s_i)$ represents the cosine similarity function in the foundation representation space $\phi$ ; $\kappa$ is the concentration parameter, and $C(\kappa)$ is a normalization constant. 
By applying this kernel to the success buffer $\mathcal{B}^+$ and the failure buffer $\mathcal{B}^-$, we can directly formulate the log-densities. By embedding the non-parametric expectation within the logarithm and separating the multiplicative constants into additive ones, we obtain:
\begin{equation}
\log p_{+}(s) = \log \left[ C(\kappa^+) \mathbb{E}_{\mathcal{B}^+} \exp\left( \kappa^+\text{sim}_\phi(s, g^+) \right) \right] = \log \mathbb{E}_{\mathcal{B}^+} \exp\left( \kappa^+\text{sim}_\phi(s, g^+) \right) + C^+
\end{equation}
\begin{equation}
\log p_{-}(s) = \log \left[ C(\kappa^-) \mathbb{E}_{\mathcal{B}^-} \exp\left( \kappa^-\text{sim}_\phi(s, g^-) \right) \right] = \log \mathbb{E}_{\mathcal{B}^-} \exp\left( \kappa^-\text{sim}_\phi(s, g^-) \right) + C^-
\end{equation}
Let LogSumExp operator $\mathbb{L}^\kappa_\mathcal{B}(X) := \frac{1}{\kappa} \log \mathbb{E}_{\mathcal{B}}[\exp(\kappa X)]$. Substituting these refined log-densities back into Eq.~\ref{eq:irl} yields the complete, closed-form formulation of the potential-based rewards in HiRE:
\begin{equation}
\Phi_{\text{HiRE}}(s) :=\frac{1}{\kappa^+}R^*(s)= \mathbb{L}^{\kappa^+}_{\mathcal{B}^+}\left(\text{sim}_\phi(s, g^+) \right) - \lambda\cdot\mathbb{L}^{\kappa^-}_{\mathcal{B}^-}\left(\text{sim}_\phi(s, g^-) \right)
\label{eq:hire_logsumexp}
\end{equation}
where $\lambda = \kappa^-/\kappa^+$ naturally emerges as the ratio of the two concentration parameters that aligns the landscapes of the success and failure manifolds. 
The additive constant shift $C^+ - C^-$ is omitted as a state-independent constant shift in the reward function as it does not alter the optimal policy or the policy gradient direction in reinforcement learning~\citep{ng1999policy}.
Crucially, the magnitude of $\lambda$ reflects the geometric constraints of tasks: precision-critical assembly tasks naturally demand $\lambda < 1$ ($\kappa^+ > \kappa^-$) because the successful manifold is extremely narrow and localized, i.e. high concentration, whereas failure modes are broad and generic. This configuration inherently shapes an asymmetric potential field, providing a steep, sharp attraction gradient toward the microscopic goal alongside a broad, gentle repulsion basin that steers the agent away from failure traps without suppressing local exploration.

Finally, to augment foundation representations with control awareness while preserving the optimality of the policy, we integrate this control-aware potential $\Phi_{\text{HiRE}}$ with standard task success indicators $R_{\text{sparse}}(s,a,s^\prime)$ (i.e. sparse reward) via potential-based reward shaping~\citep{ng1999policy}:
\begin{equation}
R_{\text{HiRE}}(s,a,s^\prime) = R_{\text{sparse}}(s,a,s^\prime) + w\left(\gamma\Phi_{\text{HiRE}}(s^\prime)-\Phi_{\text{HiRE}}(s)\right)\label{eq:hir_final}
\end{equation}
where $\gamma$ is the discount factor,
ensuring that the edited reward landscape provides dense guidance without introducing any unintended optima deviated from task objective.

\paragraph{Intuitive interpretations.}
The subtractive structure of the HiRE potential in Eq.~\ref{eq:hire_logsumexp} directly corrects the control-unawareness of raw foundation representations, whose conventional similarity-based rewards rely strictly on passive visual matching and remain unaware of physical task dynamics. 
By contrasting hindsight interaction data, HiRE partitions the representation space into three distinct geometric regimes that systematically resolve the misalignment between visual similarity and control objectives:
(1) \textbf{Success-dominated regime (correcting false negatives):} Raw foundation representations evaluate task progress purely upon visual features, routinely underestimating intermediate states that are dynamically correct yet appear visually distant from the final goal (e.g., a correct motion reaching the target object that occurs far from the ultimate placement site). HiRE elevates the potential of these states by aggregating successful histories, injecting a strong attractive gradient that rescues these false negatives from being under-rewarded due to visual disparity.
(2) \textbf{Failure-dominated regime (correcting false positives):} Conversely, semantic similarity is inherently blind to physical constraints, making it highly prone to overestimating deceptive ``trap states''---states that appear visually identical to the goal but have suffered irreversible physical failures (e.g., moving toward the placement site after dropping the small object). HiRE imposes a severe penalty on these zones, explicitly suppressing visual reward hacking that drives the robot into physical failures.
(3) \textbf{Equilibrium regime (gradient cancellation):} In equilibrium zones or vast under-covered regions, the gradients of the success and failure terms achieve a dynamic counter-balance. Mathematically, even though the potential values do not identically sum to zero due to $\lambda < 1$, the reward landscape converges to a flat constant plateau where the localized shaping gradient vanishes, i.e., $\nabla_s \Phi_{\text{HiRE}}(s) \approx \mathbf{0}$. This gradient cancellation allows HiRE to gracefully degrade to the clean sparse-reward recipe. Consequently, it prevents HiRE from issuing conflicting signals on states from blurred manifolds and ensures the robot is not overly penalized by localized failures when exploring distant, out-of-distribution states.


\subsection{RL Policy Finetuning with HiRE}
To dynamically evaluate the non-parametric densities, HiRE maintains a positive buffer $\mathcal{B}^+$ and a negative buffer $\mathcal{B}^-$ during policy finetuning. The positive buffer $\mathcal{B}^+$ consists of an offline buffer constructed from expert demonstrations, together with an online first-in-first-out (FIFO) buffer that stores observations from recent successful rollouts. The negative buffer $\mathcal{B}^-$ is another FIFO buffer, populated only with terminal states from failed rollouts to illustrate the failure manifold.
As HiRE only reshapes the intrinsic reward signal, it is agnostic to the underlying RL algorithm. We formalize this plug-and-play integration through a generalized actor-critic objective. Let $\mathcal{U}$ denote the sampling distribution from which the backbone draws training transitions $(s,a,s’)$. The critic $Q_\psi$ is updated by minimizing the shaped Bellman residual:
\begin{equation}
\mathcal{L}_\psi = \mathbb{E}_{(s, a, s^\prime) \sim \mathcal{U}} \left[  Q_\psi(s, a) - \big( R_{\text{HiRE}}(s,a,s^\prime) + \gamma \mathbb{E}_{a' \sim \pi_\theta}[Q_{\bar{\psi}}(s^\prime, a')] \big)  \right]^2
\end{equation}
where $Q_{\bar{\psi}}$ is the target network; $\bar{\psi}$ is periodically copied from $\psi$; $R_{\text{HiRE}}(s,a,s^\prime)$ in Eq.~\ref{eq:hir_final} represents our dense reward recipe that draw policies toward success and away from failure, where $\lambda\in[0,1)$ as success manifolds are usually higher concentrated than failure ones. Subsequently, the actor $\pi_\theta$ is optimized to minimize the negative estimated shaped return $Q_\psi$ alongside a flexible regularization term $\mathcal{L}_{\texttt{algo}}$ which depends on choice of RL algorithm \texttt{algo}: 
\begin{equation}
\mathcal{L}_\theta= \mathbb{E}_{s \sim \mathcal{U}, a \sim \pi_\theta} \left[ -\omega_{\texttt{algo}}(s, a) \cdot Q_\psi(s, a) +\mathcal{L}_{\texttt{algo}}\right]
\end{equation}
This abstraction elegantly accommodates various RL backbones: setting $\omega_{\texttt{algo}} = 1$ recovers standard deterministic or stochastic off-policy methods; incorporating $\omega_{\texttt{algo}}$ with clipped importance sampling constraints yields stable on-policy updates (e.g., PPO~\citep{schulman2017proximal}); and applying behavior regularization $\mathcal{L}_{\texttt{algo}}$ enables entropy-based exploration bonus for stochastic policies  (e.g., TD3~\citep{fujimoto2018addressing}, SAC~\citep{haarnoja2018soft}) or safe finetuning from the base policy (e.g., DICE-RL~\citep{sun2026prior}). In our main implementations, we integrate our reward recipe on top of DICE-RL, a stable and sample-efficient RL policy finetuning algorithm.

\section{Experiments}

We evaluate HiRE in the setting of online RL finetuning of pretrained robot policies. We adopt DINOv2~\citep{oquab2024dinov} and SigLIP~\citep{tschannen2025siglip} as foundation encoders, and DICE-RL~\citep{sun2026prior}, a stable and sample-efficient finetuning framework for generative behavior cloning policies, as policy learning backbone. 
Building on such a competitive backbone lets us cleanly isolate the contribution of the reward signal, and ask whether a better reward can push an already strong finetuning pipeline even further.
Concretely, our experiments are designed around three questions: Can HiRE improve an RL finetuning backbone with dense and correct guidance? How does HiRE compare with learned reward adaptation, pretrained robotic reward models, and training-free VLM rewards? How does HiRE dynamically shape reward landscape that eventually lifts policy finetuning ceiling? Implementation details, additional comparative, hyperparameter sensitivity and ablation results are elaborated on in App.~\ref{app:addition}.

\subsection{Experimental Setup}
\textbf{(1) Benchmarks and tasks.}
We evaluate HiRE on both simulation and real-robot manipulation tasks. 
We use RoboMimic~\citep{mandlekar2022matters} and MimicGen~\citep{mandlekar2023mimicgen} simulation benchmarks,
and test on two different manipulation tasks with I2RT YAM robot arms for real-robot experiments, detailed in App.~\ref{app:benchmark} and \ref{app:real} respectively.
Together, these tasks span prehensile, long-horizon, contact-rich, and high-precision manipulation, with several requiring high-precision insertion and assembly. For each task, we first train generative behavior cloning (BC) policies from offline demonstrations on two benchmarks respectively, achieving base policies with success rates of $18.0\%$ on \texttt{Stack\_Three}, $6.0\%$ on \texttt{Three\_Piece\_Assembly}, $1.0\%$ on \texttt{Threading}, $2.0\%$ on \texttt{Tool\_Hang}, and $20.0\%$ on both hardware tasks, i.e., \texttt{Bottle\_Pick-and-Place} and \texttt{Towers\_of\_Hanoi}, and then finetune them with DICE-RL. 
\textbf{(2) Reward baselines.}
We compare HiRE with four reward baselines. 
\textbf{Sparse reward} uses binary task success signals. 
It is easy to obtain in both simulation and real-world settings but provides no intermediate feedback before task completion, which falls short on long-horizon multi-stage tasks. 
\textbf{RoboMeter}~\citep{liang2026robometer} (with \texttt{RoboMeter-4B} configuration) is a general-purpose robotic dense reward model trained with frame-level progress supervision and trajectory-level preference supervision. 
It learns from trajectory comparisons among expert, suboptimal, and failed rollouts, with large-scale task-specific human labeling. 
\textbf{TOPReward}~\citep{chen2026topreward} (with \texttt{Qwen3-VL-8B-Instruct} configuration) is a training-free robotic reward method that estimates task progress from a pretrained VLM.
It scores task-completion token probabilities conditioned on trajectory prefixes and task instructions, without task-specific reward-model training or human progress annotations.
\textbf{GCR}~\citep{biza2025robot} learns dense rewards by combining implicit value learning with a goal-contrastive objective that distinguishes successful and failed trajectories.
It provides a learned reward-adaptation baseline for comparison with HiRE's training-free reward editing.


\begin{figure*}[t]
\centering
\includegraphics[width=\linewidth]{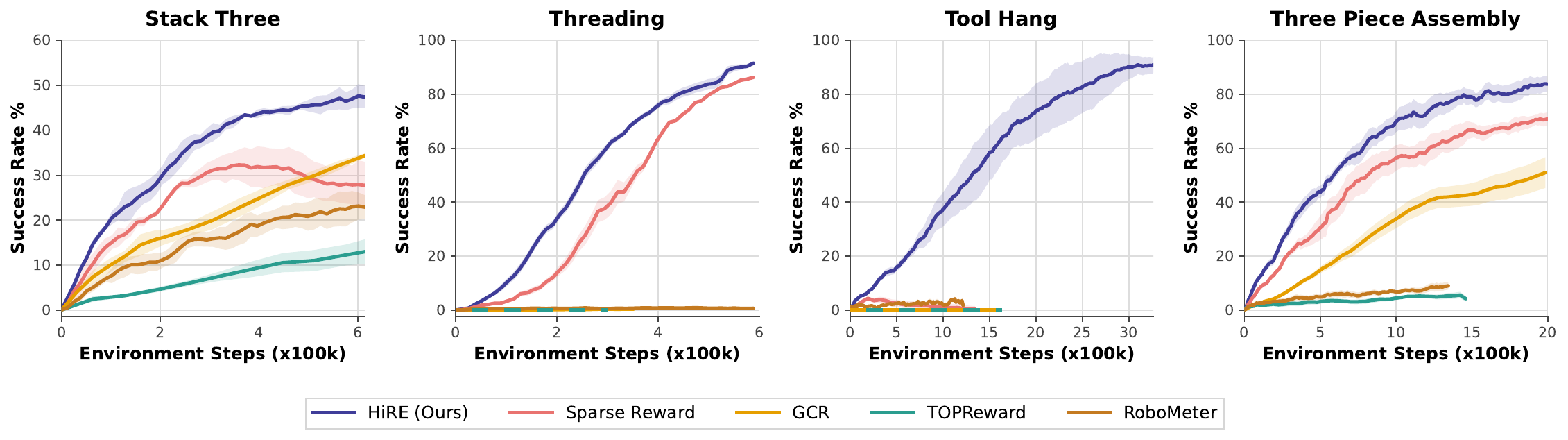}
\caption{
\textbf{Comparisons on RoboMimic and MimicGen tasks.}
Success rate versus online environment steps for different reward recipes integrated with DICE-RL policy learning backbone. Curves are averaged over three random seeds, with shaded regions denoting variability across seeds.}
\label{fig:main_curves}
\vspace{-10pt}
\end{figure*}

\subsection{Benchmark Comparative Results}

\paragraph{Long-horizon, multi-stage tasks {\normalfont(\texttt{Stack\_Three}, \texttt{Three\_Piece\_Assembly}).}} For these long-horizon tasks, \textbf{HiRE lifts the final success rate substantially above the sparse reward baseline} as shown in Fig.~\ref{fig:main_curves}. On \texttt{Stack\_Three}, sparse-reward policy finetuning initially improves but later declines to around $30\%$, while HiRE continues to improve and reaches a success rate of roughly $50\%$. On \texttt{Three\_Piece\_Assembly}, sparse-reward finetuning reaches a final success rate of around $70\%$, while HiRE climbs to over $80\%$. 
Since both tasks require several ordered sub-stages before terminal success, sparse rewards provide limited credit assignment within available interaction budgets. In contrast, HiRE supplies dense progress feedback, leading to higher performance.

\paragraph{Contact-rich and high-precision tasks {\normalfont(\texttt{Tool\_Hang}, \texttt{Threading}).}} For contact-rich and precision tasks, HiRE's advantage appears in two related but distinct aspects. On \texttt{Tool\_Hang}, sparse-reward finetuning fails completely despite non-zero base policy performance, whereas HiRE steadily improves to a success rate of around $95\%$ as shown in Fig.~\ref{fig:main_curves}. 
HiRE provides dense feedback on intermediate progress, helping the policy learn useful manipulation behaviors even when task successes are rare. On \texttt{Threading}, though the sparse-reward recipe eventually reaches a comparable final success rate, \textbf{HiRE always reaches the same performance with markedly fewer online interactions}, showing clear gains in sample efficiency.

\paragraph{Comparison with other reward baselines.}
The above results show that HiRE raises performance when sparse rewards fail
to capture intermediate progress and at least improves sample efficiency when sparse
supervision is eventually sufficient. We further compare HiRE with pretrained
reward models and methods that adapt rewards online.
The pretrained robotic reward model \textbf{RoboMeter} and the VLM-based reward
method \textbf{TOPReward} exhibit inconsistent performance across tasks and achieve
substantially lower final success rates than HiRE, suggesting limited
generalization of their reward signals to these unseen tasks and scenarios.
\textbf{GCR} also adapts rewards using online failure experience, but requires
gradient-based reward-model updates. On the multi-stage tasks (\texttt{Stack\_Three}, \texttt{Three\_Piece\_Assembly}), GCR requires roughly $3$
times as many interactions as HiRE to reach $30\%$ success. On the high-precision tasks (\texttt{Tool\_Hang}, \texttt{Threading}), it fails to make substantial progress. Together, these comparisons highlight HiRE's sample
efficiency and performance consistency across tasks, achieved through its training-free dense reward recipe.




\begin{figure*}[t]
\centering
\includegraphics[width=\linewidth]{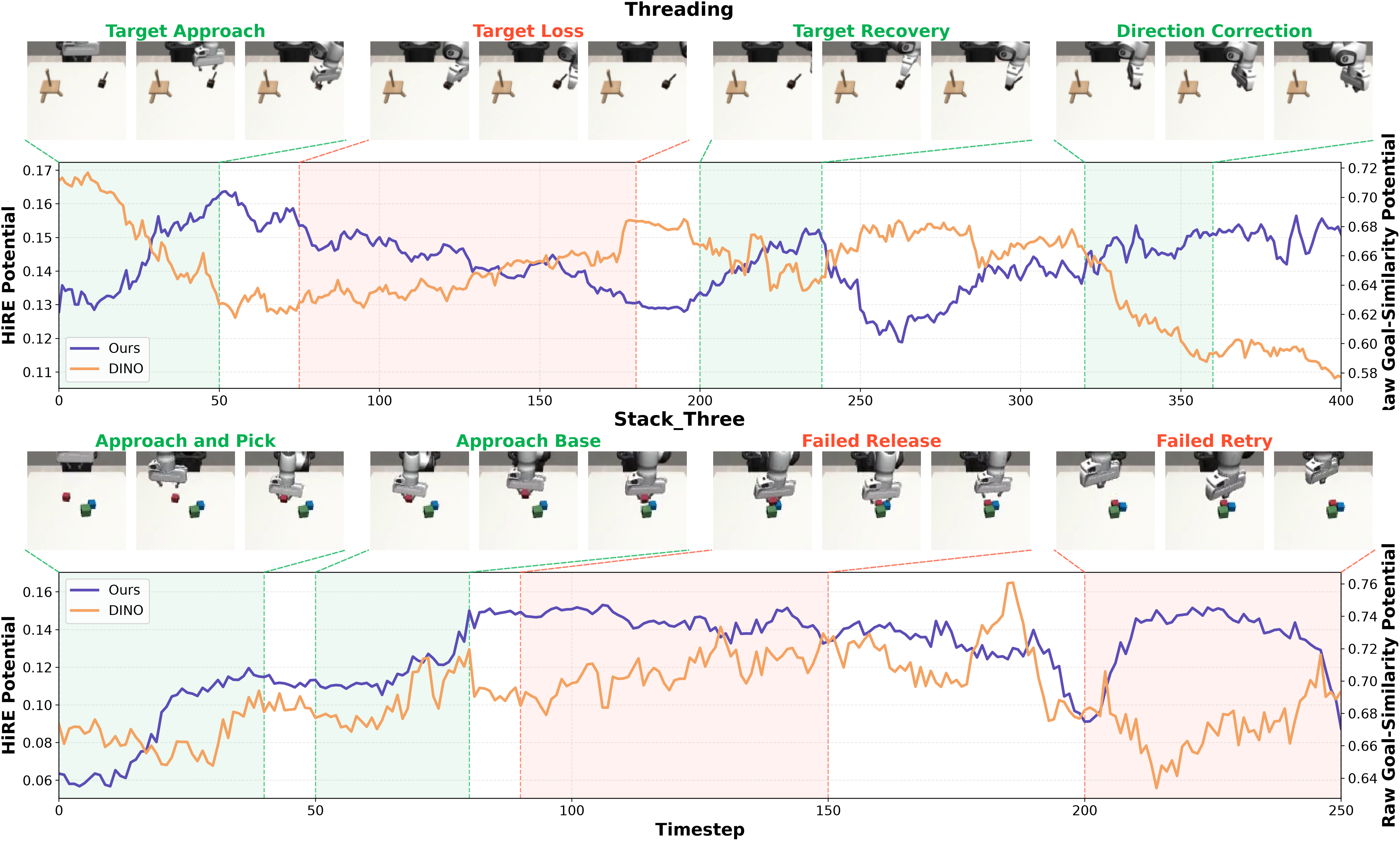}
\caption{
    \textbf{HiRE reflects task progress better than na\"ive visual similarity-based rewards.}
    We visualize reward scores along two representative trajectories, with selected segments highlighted and shown as thumbnails above the plot.
    The shaded regions correspond to key behavioral phases, including target approach, target loss, target recovery, and direction correction.
    While DINO similarity-based potentials mainly reflect visual resemblance to the ultimate goal frame, the HiRE potential remains faithful to reflect task progress and penalizes visually plausible but failure-prone states.
}
\label{fig:reward_vis}
\vspace{-15pt}
\end{figure*}


\subsection{Real-Robot Comparative Results}

\begin{figure}[ht]
    \centering
    \includegraphics[width=\linewidth]{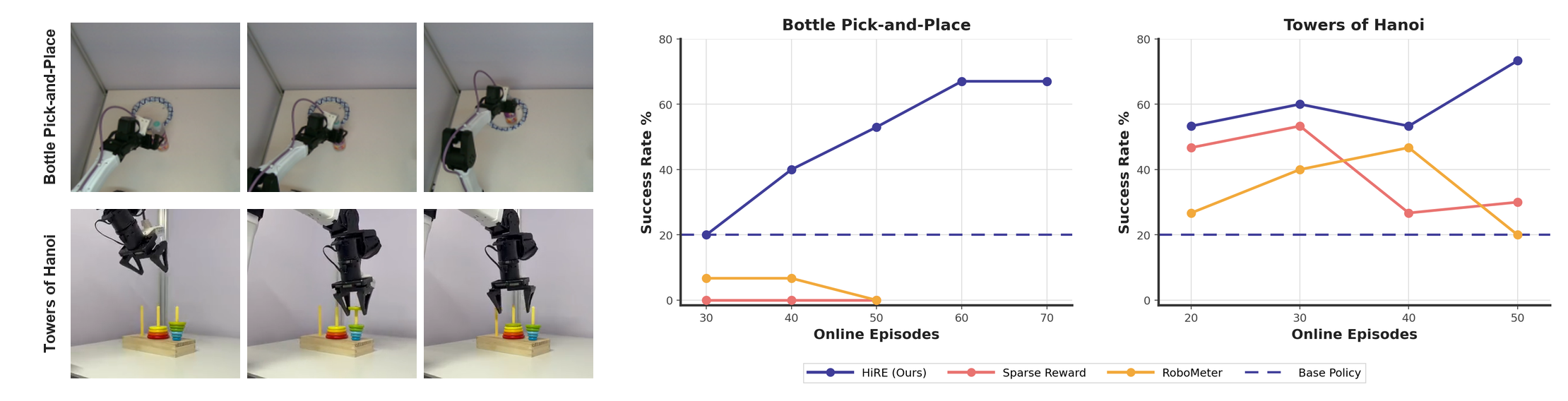}
    \caption{
        \textbf{Real-robot evaluation.}
        \textbf{Left:} Representative successful HiRE rollouts for \texttt{Bottle\_Pick-and-Place} (top) and \texttt{Towers\_of\_Hanoi} (bottom).
        \textbf{Right:} Success rates over online fine-tuning episodes.
        Starting from the same base policy ($20\%$), HiRE reaches $67\%$ on \texttt{Bottle\_Pick-and-Place} and $73\%$ on \texttt{Towers\_of\_Hanoi}, consistently outperforming sparse-reward and RoboMeter baselines.
    }
    \label{fig:real_curve}
    \vspace{-10pt}
\end{figure}


We evaluate each method on two real-robot tasks, comparing their success rates per checkpoint in Fig.~\ref{fig:real_curve}. 
On \texttt{Bottle\_Pick-and-Place}, HiRE steadily guides the policy toward better performance, boosting the success rate from $20\%$ to $67\%$, whereas the sparse-reward and Robometer baselines nearly fail entirely in the early training phases. 
Detailed performance across all six initial positions is further reported in Tab.~\ref{tab:real_robot_results}. 
On \texttt{Towers\_of\_Hanoi}, HiRE instantly boosts the success rate to 53.3\% with only 20 episodes, eventually reaching a peak of \textbf{73.3\%} (3.67$\times$ base policy). In contrast, the baselines achieve substantially lower peak performances (Sparse: 53.3\% at 30 episodes; Robometer: 46.7\% at 40 episodes), requiring more interactions to reach inferior caps.
While HiRE maintains steady performance gains, both baselines suffer from severe degradation during fine-tuning (e.g., Sparse drops to 30.0\% and Robometer to 20.0\% at 50 episodes) due to reward hacking or noisy reward estimates. 
In conclusion, HiRE achieves superior sample efficiency, stable improvement, and higher policy performance ceiling on both hardware tasks.

\vspace{-2pt}
\begin{wrapfigure}{r}{0.4\linewidth}
    \centering
    \vspace{-10pt} 
    \includegraphics[width=\linewidth]{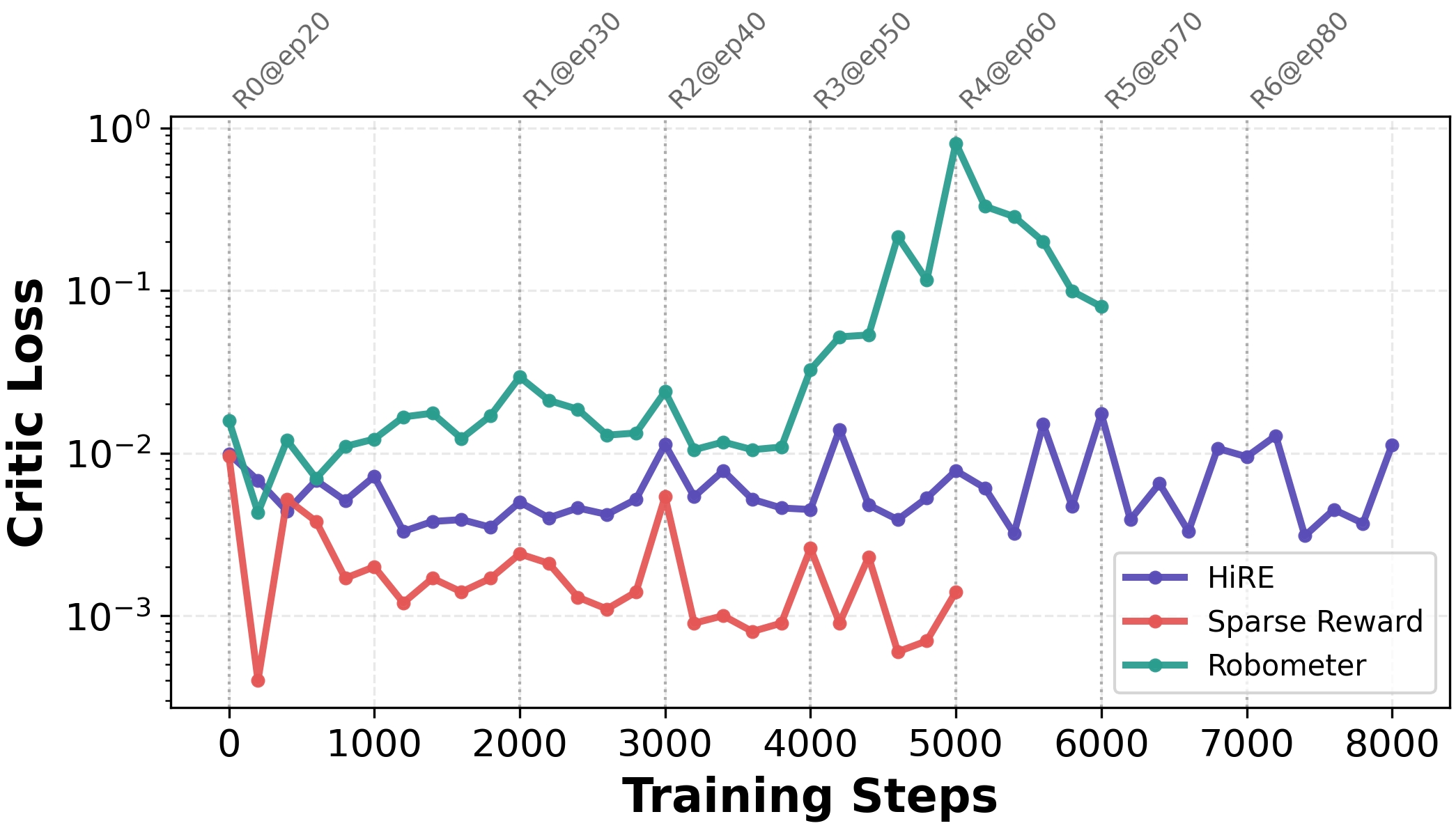}
    \caption{
        \textbf{Critic loss.}
        We compare critic loss between HiRE and baselines during policy finetuning.
        HiRE maintains non-trivial critic updates, while sparse reward yields significantly lower critic loss due to its sparse signals, and Robometer probably produces inconsistent rewards that are hard for the critic to converge.
    }
    \label{fig:critic_loss_sparse_vs_hire}
    \vspace{-10pt} 
\end{wrapfigure}

This significant contrast stems from HiRE's ability to extract dense optimality signals from the positive and negative buffers, which efficiently activates and steers the base policy toward successful cases. 
Conversely, sparse rewards fail to provide sufficient optimization gradients. 
As illustrated by the critic loss comparison during online training on \texttt{Bottle\_Pick-and-Place} in Fig.~\ref{fig:critic_loss_sparse_vs_hire}, sparse rewards yield a prematurely low critic loss due to the severe sparsity of reward signals, leading to value function collapse. 
In contrast, Robometer induces a non-converged critic loss, probably because it introduces extrinsic rewards that are not robust with OOD observations.
HiRE maintains non-trivial critic updates throughout the finetuning process, demonstrating that its edited potential field continuously delivers informative feedback to sustain effective policy learning.

\subsection{Visualizations and Analyses}\label{sec:vis}
\paragraph{HiRE endows foundation vision representations with control-awareness.} In Fig.~\ref{fig:reward_vis}, we compare the raw DINO goal-similarity potential $\Phi_\text{raw}(s)=\text{sim}_\phi(s, g^+_\star)$ commonly adopted by existing representation-based reward learning methods~\citep{ma2022vip,ma2023liv,li2024decisionnce}, against our edited potential $\Phi_{\text{HiRE}}(s)$ in Eq.~\ref{eq:hire_logsumexp} on \texttt{Threading} and \texttt{Stack\_Three} trajectories. Taking \texttt{Threading} subplot as an example, the robot correctly approaches the target object during the early execution phase (e.g. Timestep 0-50). 
However, because this leads to observations visually distinct from the ultimate goal frame $g^+_\star$, $\Phi_\text{raw}(s)$ severely degrades, potentially causing optimization misdirection. 
In sharp contrast, $\Phi_{\text{HiRE}}(s)$ successfully aggregates historical success data to steadily elevate the reward, reinforcing this critical task progress and correcting such false negatives of $\Phi_\text{raw}(s)$ caused by visual disparity.
Conversely, a more catastrophic visual artifact occurs during the bad navigation phases (e.g. Timestep 75-180), where the robot loses its target and tracks overshoots, causing its leaving the camera view. Paradoxically, $\Phi_\text{raw}(s)$ exhibits a deceptive increase, falsely interpreting the empty background as a closer match to $g^+_\star$. 
These highlight a fundamental flaw in raw rewards derived from foundation vision representations: $\Phi_\text{raw}(s)$ is inherently vulnerable to both \emph{false negatives} (under-rewarding visually incorrect yet beneficial behaviors) and \emph{false positives} (over-rewarding visually correct yet improper behaviors). By attracting towards success and repelling from failure manifolds, HiRE systematically unmasks these visual illusions and grounds control-awareness into rewards.



\begin{figure*}[t]
    \centering
    \includegraphics[width=\textwidth]{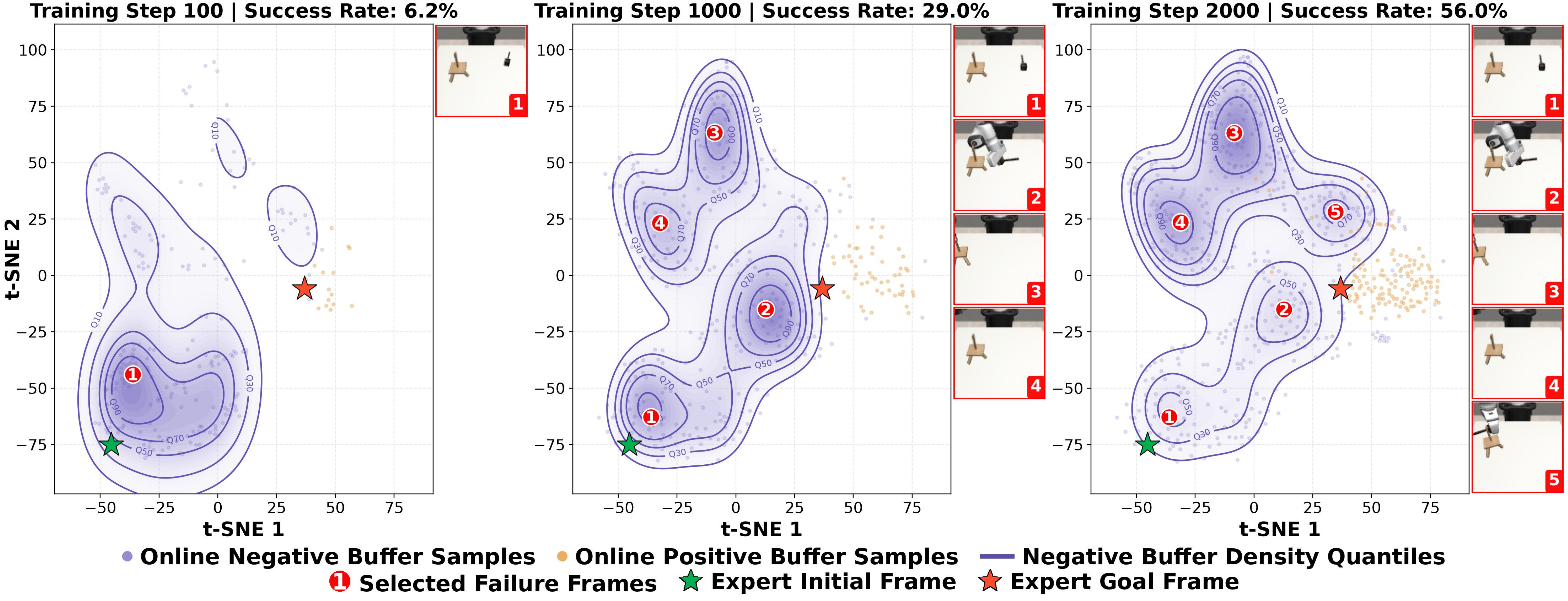}
    \caption{
        \textbf{The negative buffer captures distinct failure behaviors during policy finetuning.}
        We visualize positive and negative buffer samples in a shared t-SNE space across training steps.
        Online positive buffer samples correspond to successful rollouts whereas negative ones are the terminal frames of failure rollouts.
        Density quantile contours highlight high-density regions of the negative-buffer distribution, where selected failure frames are sampled as representative examples.
        Expert initial and goal frames are plotted as reference markers.
    }
    \label{fig:buffer_samples_scatter_plot}
    \vspace{-18pt}
\end{figure*}

\paragraph{HiRE co-adapts the reward manifold upon policy finetuning performance.} In Fig.~\ref{fig:buffer_samples_scatter_plot}, we analyze the joint t-SNE embedding space of both online positive and negative buffer samples across training steps. The empirical density quantile contours reveal two profound insights into dynamical reward shaping nature of HiRE.
First, online negative buffer dynamically captures stage-progressive failure modes while preventing over-penalization. At the beginning, the negative samples concentrates around the initial state (Region 1). As the policy performance improves, the buffer removes these early, trivial failures and segments into multiple localized clusters (Regions 2-5) that encapsulate harder, later-stage failures. HiRE thus avoids global over-penalization and maintains a targeted shaping to only suppress localized failures along training.
Second, the positive distribution remains highly concentrated whereas the negative one is much wider and scattered. This empirical discrepancy confirms assigning a larger concentration factor $\kappa^+$ than to $\kappa^-$, which we discuss in Sec.~\ref{hire}.


\section{Conclusion and Limitations}\label{conclusion}
HiRE addresses the reward bottleneck in RL policy finetuning by endowing foundation-representation similarity with control-awareness. Without training reward models or using human labels, HiRE edits rewards online through hindsight evidence from successful and failed rollouts, suppressing deceptive trap states while preserving genuine task progress. Across RoboMimic, MimicGen, and real-robot platforms, this lightweight calibration consistently improves both final success rate and sample efficiency over other reward methods, while remaining agnostic to the encoder and finetuning algorithm. Overall, HiRE offers a practical and scalable recipe that helps existing RL finetuning pipelines unlock their full potential.
\textbf{Limitations and Future Work.} 
While success and failure signals offer highly accessible and clean feedback, they overlook intermediate, control-centric information. Future extensions could incorporate human guidance to make interaction data more control-aware, thereby shaping a more nuanced reward landscape.
Another promising direction is to develop more sophisticated buffer construction strategies that account for suboptimal segments within successful trajectories, such as segments involving retries or recovery behaviors of varying degrees.

\section*{Acknowledgement}
We express our sincere gratitude to Neural Motion and the Tianqiao \& Chrissy Chen Institute for providing essential computational resources for this work. In particular, we deeply appreciate Neural Motion for supplying critical GPU compute in the final month prior to submission. We also thank Zhanyi Sun and Qianzhong Chen from Stanford University for their valuable insights and technical guidance on implementation details.

\bibliography{example}  

@article{schulman2017proximal,
	title={Proximal policy optimization algorithms},
	author={Schulman, John and Wolski, Filip and Dhariwal, Prafulla and Radford, Alec and Klimov, Oleg},
	journal={arXiv preprint arXiv:1707.06347},
	year={2017}
}

@inproceedings{fujimoto2018addressing,
	title={Addressing Function Approximation Error in Actor-Critic Methods},
	author={Fujimoto, Scott and Hoof, Herke and Meger, David},
	booktitle={International Conference on Machine Learning},
	pages={1587--1596},
	year={2018}
}

@inproceedings{haarnoja2018soft,
	title={Soft Actor-Critic: Off-Policy Maximum Entropy Deep Reinforcement Learning with a Stochastic Actor},
	author={Haarnoja, Tuomas and Zhou, Aurick and Abbeel, Pieter and Levine, Sergey},
	booktitle={International Conference on Machine Learning},
	pages={1861--1870},
	year={2018}
}

@book{boyd2004convex,
	title={Convex optimization},
	author={Boyd, Stephen and Boyd, Stephen P and Vandenberghe, Lieven},
	year={2004},
	publisher={Cambridge university press}
}

@article{sutton1998introduction,
  title={Introduction to reinforcement learning},
  author={Sutton, Richard S and Barto, Andrew G and others},
  year={1998},
  publisher={MIT press Cambridge}
}

@article{ball2023efficient,
  title={Efficient online reinforcement learning with offline data},
  author={Ball, Philip J and Smith, Laura and Kostrikov, Ilya and Levine, Sergey},
  journal={arXiv preprint arXiv:2302.02948},
  year={2023}
}

@inproceedings{ma2023liv,
  title={Liv: Language-image representations and rewards for robotic control},
  author={Ma, Yecheng Jason and Kumar, Vikash and Zhang, Amy and Bastani, Osbert and Jayaraman, Dinesh},
  booktitle={International Conference on Machine Learning},
  pages={23301--23320},
  year={2023},
  organization={PMLR}
}

@article{ma2022vip,
  title={Vip: Towards universal visual reward and representation via value-implicit pre-training},
  author={Ma, Yecheng Jason and Sodhani, Shagun and Jayaraman, Dinesh and Bastani, Osbert and Kumar, Vikash and Zhang, Amy},
  journal={arXiv preprint arXiv:2210.00030},
  year={2022}
}

@inproceedings{biza2025robot,
  title={On-Robot Reinforcement Learning with Goal-Contrastive Rewards},
  author={Biza, Ondrej and Weng, Thomas and Sun, Lingfeng and Schmeckpeper, Karl and Kelestemur, Tarik and Ma, Yecheng Jason and Platt, Robert and van de Meent, Jan-Willem and Wong, Lawson LS},
  booktitle={2025 IEEE International Conference on Robotics and Automation (ICRA)},
  pages={4797--4805},
  year={2025},
  organization={IEEE}
}

@article{sun2026prior,
  title={From Prior to Pro: Efficient Skill Mastery via Distribution Contractive RL Finetuning},
  author={Sun, Zhanyi and Song, Shuran},
  journal={arXiv preprint arXiv:2603.10263},
  year={2026}
}

@inproceedings{mandlekar2022matters,
  title={What Matters in Learning from Offline Human Demonstrations for Robot Manipulation},
  author={Mandlekar, Ajay and Xu, Danfei and Wong, Josiah and Nasiriany, Soroush and Wang, Chen and Kulkarni, Rohun and Fei-Fei, Li and Savarese, Silvio and Zhu, Yuke and Mart{\'\i}n-Mart{\'\i}n, Roberto},
  booktitle={Conference on Robot Learning},
  pages={1678--1690},
  year={2022},
  organization={PMLR}
}

@inproceedings{mandlekar2023mimicgen,
  title={MimicGen: A Data Generation System for Scalable Robot Learning using Human Demonstrations},
  author={Mandlekar, Ajay and Nasiriany, Soroush and Wen, Bowen and Akinola, Iretiayo and Narang, Yashraj and Fan, Linxi and Zhu, Yuke and Fox, Dieter},
  booktitle={Conference on Robot Learning},
  pages={1820--1864},
  year={2023},
  organization={PMLR}
}

@inproceedings{wagenmaker2025steering,
  title={Steering Your Diffusion Policy with Latent Space Reinforcement Learning},
  author={Wagenmaker, Andrew and Zhang, Yunchu and Nakamoto, Mitsuhiko and Park, Seohong and Yagoub, Waleed and Nagabandi, Anusha and Gupta, Abhishek and Levine, Sergey},
  booktitle={Conference on Robot Learning},
  pages={258--282},
  year={2025},
  organization={PMLR}
}

@article{krack2026rewarding,
  title={Rewarding DINO: Predicting Dense Rewards with Vision Foundation Models},
  author={Krack, Pierre and J{\"u}lg, Tobias and Burgard, Wolfram and Walter, Florian},
  journal={arXiv preprint arXiv:2603.16978},
  year={2026}
}

@article{xu2026rl,
  title={RL Token: Bootstrapping Online RL with Vision-Language-Action Models},
  author={Xu, Charles and Springenberg, Jost Tobias and Equi, Michael and Amin, Ali and Esmail, Adnan and Levine, Sergey and Ke, Liyiming},
  journal={arXiv preprint arXiv:2604.23073},
  year={2026}
}

@article{ankile2025residual,
  title={Residual off-policy rl for finetuning behavior cloning policies},
  author={Ankile, Lars and Jiang, Zhenyu and Duan, Rocky and Shi, Guanya and Abbeel, Pieter and Nagabandi, Anusha},
  journal={arXiv preprint arXiv:2509.19301},
  year={2025}
}

@inproceedings{
yuan2025policy,
title={Policy Decorator: Model-Agnostic Online Refinement for Large Policy Model},
author={Xiu Yuan and Tongzhou Mu and Stone Tao and Yunhao Fang and Mengke Zhang and Hao Su},
booktitle={The Thirteenth International Conference on Learning Representations},
year={2025},
url={https://openreview.net/forum?id=e5jGTEiJMT}
}

@inproceedings{
chen2026sarm,
title={SARM: Stage-Aware Reward Modeling for Long Horizon Robot Manipulation},
author={Qianzhong Chen and Justin Yu and Mac Schwager and Pieter Abbeel and Fred Shentu and Philipp Wu},
booktitle={The Fourteenth International Conference on Learning Representations},
year={2026},
url={https://openreview.net/forum?id=aemqAxScl9}
}

@inproceedings{yang2024rank2reward,
  title={Rank2reward: Learning shaped reward functions from passive video},
  author={Yang, Daniel and Tjia, Davin and Berg, Jacob and Damen, Dima and Agrawal, Pulkit and Gupta, Abhishek},
  booktitle={2024 IEEE International Conference on Robotics and Automation (ICRA)},
  pages={2806--2813},
  year={2024},
  organization={IEEE}
}

@inproceedings{nair2023r3m,
  title={R3M: A Universal Visual Representation for Robot Manipulation},
  author={Nair, Suraj and Rajeswaran, Aravind and Kumar, Vikash and Finn, Chelsea and Gupta, Abhinav},
  booktitle={Conference on Robot Learning},
  pages={892--909},
  year={2023},
  organization={PMLR}
}

@inproceedings{li2024decisionnce,
  title={DecisionNCE: embodied multimodal representations via implicit preference learning},
  author={Li, Jianxiong and Zheng, Jinliang and Zheng, Yinan and Mao, Liyuan and Hu, Xiao and Cheng, Sijie and Niu, Haoyi and Liu, Jihao and Liu, Yu and Liu, Jingjing and others},
  booktitle={Proceedings of the 41st International Conference on Machine Learning},
  pages={29461--29488},
  year={2024}
}

@article{
oquab2024dinov,
title={{DINO}v2: Learning Robust Visual Features without Supervision},
author={Maxime Oquab and Timoth{\'e}e Darcet and Th{\'e}o Moutakanni and Huy V. Vo and Marc Szafraniec and Vasil Khalidov and Pierre Fernandez and Daniel HAZIZA and Francisco Massa and Alaaeldin El-Nouby and Mido Assran and Nicolas Ballas and Wojciech Galuba and Russell Howes and Po-Yao Huang and Shang-Wen Li and Ishan Misra and Michael Rabbat and Vasu Sharma and Gabriel Synnaeve and Hu Xu and Herve Jegou and Julien Mairal and Patrick Labatut and Armand Joulin and Piotr Bojanowski},
journal={Transactions on Machine Learning Research},
issn={2835-8856},
year={2024},
url={https://openreview.net/forum?id=a68SUt6zFt},
note={Featured Certification}
}

@article{tschannen2025siglip,
  title={Siglip 2: Multilingual vision-language encoders with improved semantic understanding, localization, and dense features},
  author={Tschannen, Michael and Gritsenko, Alexey and Wang, Xiao and Naeem, Muhammad Ferjad and Alabdulmohsin, Ibrahim and Parthasarathy, Nikhil and Evans, Talfan and Beyer, Lucas and Xia, Ye and Mustafa, Basil and others},
  journal={arXiv preprint arXiv:2502.14786},
  year={2025}
}

@inproceedings{
mu2024drs,
title={DrS: Learning Reusable Dense Rewards for Multi-Stage Tasks},
author={Tongzhou Mu and Minghua Liu and Hao Su},
booktitle={The Twelfth International Conference on Learning Representations},
year={2024},
url={https://openreview.net/forum?id=6CZ50WgfCG}
}

@inproceedings{ng1999policy,
  title={Policy Invariance Under Reward Transformations: Theory and Application to Reward Shaping},
  author={Ng, Andrew Y and Harada, Daishi and Russell, Stuart J},
  booktitle={Proceedings of the Sixteenth International Conference on Machine Learning},
  pages={278--287},
  year={1999}
}

@inproceedings{ziebart2008maximum,
  title={Maximum entropy inverse reinforcement learning.},
  author={Ziebart, Brian D and Maas, Andrew L and Bagnell, J Andrew and Dey, Anind K and others},
  booktitle={Aaai},
  volume={8},
  pages={1433--1438},
  year={2008},
  organization={Chicago, IL, USA}
}

@book{mardia2009directional,
  title={Directional statistics},
  author={Mardia, Kanti V and Jupp, Peter E},
  year={2009},
  publisher={John Wiley \& Sons}
}

@article{intelligence2025pi,
  title={$\pi_{0.6}^*$: a VLA That Learns From Experience},
  author={Intelligence, Physical and Amin, Ali and Aniceto, Raichelle and Balakrishna, Ashwin and Black, Kevin and Conley, Ken and Connors, Grace and Darpinian, James and Dhabalia, Karan and DiCarlo, Jared and others},
  journal={arXiv preprint arXiv:2511.14759},
  year={2025}
}

@article{intelligence2025pi_,
  title={$\pi_{0.5}$: a Vision-Language-Action Model with Open-World Generalization},
  author={Intelligence, Physical and Black, Kevin and Brown, Noah and Darpinian, James and Dhabalia, Karan and Driess, Danny and Esmail, Adnan and Equi, Michael and Finn, Chelsea and Fusai, Niccolo and others},
  journal={arXiv preprint arXiv:2504.16054},
  year={2025}
}

@inproceedings{liang2026robometer,
  title     = {Robometer: Scaling General-Purpose Robotic Reward Models via Trajectory Comparisons},
  author={Anthony Liang and Yigit Korkmaz and Jiahui Zhang and Minyoung Hwang and Abrar Anwar and Sidhant Kaushik and Aditya Shah and Alex S. Huang and Luke Zettlemoyer and Dieter Fox and Yu Xiang and Anqi Li and Andreea Bobu and Abhishek Gupta and Stephen Tu and Erdem Biyik and Jesse Zhang},
  year={2026},
  booktitle={Robotics: Science and Systems 2026},
}

@inproceedings{rocamonde2024vision,
  title={Vision-language models are zero-shot reward models for reinforcement learning},
  author={Rocamonde, Juan and Montesinos, Victoriano and Nava, Elvis and Perez, Ethan and Lindner, David},
  booktitle={International Conference on Learning Representations},
  volume={2024},
  pages={28446--28463},
  year={2024}
}

@article{lee2026roboreward,
  title={RoboReward: General-Purpose Vision-Language Reward Models for Robotics},
  author={Lee, Tony and Wagenmaker, Andrew and Pertsch, Karl and Liang, Percy and Levine, Sergey and Finn, Chelsea},
  journal={arXiv preprint arXiv:2601.00675},
  year={2026}
}

@article{tan2025robo,
  title={Robo-Dopamine: General Process Reward Modeling for High-Precision Robotic Manipulation},
  author={Tan, Huajie and Chen, Sixiang and Xu, Yijie and Wang, Zixiao and Ji, Yuheng and Chi, Cheng and Lyu, Yaoxu and Zhao, Zhongxia and Chen, Xiansheng and Co, Peterson and others},
  journal={arXiv preprint arXiv:2512.23703},
  year={2025}
}

@article{rolnick2019experience,
  title={Experience replay for continual learning},
  author={Rolnick, David and Ahuja, Arun and Schwarz, Jonathan and Lillicrap, Timothy and Wayne, Gregory},
  journal={Advances in neural information processing systems},
  volume={32},
  year={2019}
}

@article{chen2025pi_,
  title={$\pi\_{RL}$: Online RL Fine-tuning for Flow-based Vision-Language-Action Models},
  author={Chen, Kang and Liu, Zhihao and Zhang, Tonghe and Guo, Zhen and Xu, Si and Lin, Hao and Zang, Hongzhi and Zhang, Quanlu and Yu, Zhaofei and Fan, Guoliang and others},
  journal={arXiv preprint arXiv:2510.25889},
  year={2025}
}

@inproceedings{radford2021learning,
  title={Learning transferable visual models from natural language supervision},
  author={Radford, Alec and Kim, Jong Wook and Hallacy, Chris and Ramesh, Aditya and Goh, Gabriel and Agarwal, Sandhini and Sastry, Girish and Askell, Amanda and Mishkin, Pamela and Clark, Jack and others},
  booktitle={International conference on machine learning},
  pages={8748--8763},
  year={2021},
  organization={PmLR}
}

@article{ho2016generative,
  title={Generative adversarial imitation learning},
  author={Ho, Jonathan and Ermon, Stefano},
  journal={Advances in neural information processing systems},
  volume={29},
  year={2016}
}

@inproceedings{
fu2018learning,
title={Learning Robust Rewards with Adverserial Inverse Reinforcement Learning},
author={Justin Fu and Katie Luo and Sergey Levine},
booktitle={International Conference on Learning Representations},
year={2018},
url={https://openreview.net/forum?id=rkHywl-A-},
}

@article{robosuite2020,
title={robosuite: A Modular Simulation Framework and Benchmark for Robot Learning},
author={Zhu, Yuke and Wong, Josiah and Mandlekar, Ajay and Mart{\'i}n-Mart{\'i}n, Roberto and Joshi, Abhishek and Nasiriany, Soroush and Zhu, Yifeng},
journal={arXiv preprint arXiv:2009.12293},
year={2020},
}

@inproceedings{chi2023diffusion_policy,
title={Diffusion Policy: Visuomotor Policy Learning via Action Diffusion},
author={Chi, Cheng and Feng, Siyuan and Du, Yilun and Xu, Zhenjia and Cousineau, Eric and Burchfiel, Benjamin and Song, Shuran},
booktitle={Proceedings of Robotics: Science and Systems (RSS)},
year={2023},
}

@inproceedings{song2020ddim,
title={Denoising Diffusion Implicit Models},
author={Song, Jiaming and Meng, Chenlin and Ermon, Stefano},
booktitle={International Conference on Learning Representations (ICLR)},
year={2021},
}

@article{chen2026topreward,
  title={Topreward: Token probabilities as hidden zero-shot rewards for robotics},
  author={Chen, Shirui and Harrison, Cole and Lee, Ying-Chun and Yang, Angela Jin and Ren, Zhongzheng and Ratliff, Lillian J and Duan, Jiafei and Fox, Dieter and Krishna, Ranjay},
  journal={arXiv preprint arXiv:2602.19313},
  year={2026}
}

@InProceedings{nakamoto2025steering,
  title = 	 {Steering Your Generalists: Improving Robotic Foundation Models via Value Guidance},
  author =       {Nakamoto, Mitsuhiko and Mees, Oier and Kumar, Aviral and Levine, Sergey},
  booktitle = 	 {Proceedings of The 8th Conference on Robot Learning},
  pages = 	 {4996--5013},
  year = 	 {2025},
  editor = 	 {Agrawal, Pulkit and Kroemer, Oliver and Burgard, Wolfram},
  volume = 	 {270},
  series = 	 {Proceedings of Machine Learning Research},
  month = 	 {06--09 Nov},
  publisher =    {PMLR},
  url = 	 {https://proceedings.mlr.press/v270/nakamoto25a.html}
}

\newpage
\appendix
\newpage
\appendix
\section*{Appendix}

\section{Implementation Details}
This section provides the implementation details for the main experimental
comparison in Fig.~\ref{fig:main_curves}. We first describe the
common setup, and then detail the reward implementation for HiRE and all the baselines in
Section~\ref{sec:impl_reward_baselines}.

\subsection{Benchmark and Policy Setup}\label{app:benchmark}
\paragraph{Benchmarks.}
We use the standard RoboMimic~\citep{mandlekar2022matters} and
MimicGen~\citep{mandlekar2023mimicgen} simulation environments built on
Robosuite v1.4.1~\citep{robosuite2020}. The main experiments include
\texttt{Tool\_Hang} from RoboMimic, \texttt{Threading}, \texttt{Stack\_Three}, and
\texttt{Three\_Piece\_Assembly} from MimicGen. All environments use an \texttt{OSC\_POSE}
controller at 20\,Hz and a 7-D action space, consisting of 6-DoF end-effector
pose deltas and a gripper command. Task completion is evaluated with the
environment-provided binary success signal.
Since the simulation environments provide no failure detection mechanism, an episode is labeled as failed only if it reaches a sufficiently long timeout horizon without completing the task. Reaching this limit implies the robot either failed outright or got stuck in a local trap state, both of which are undesirable behaviors that warrant penalization. 
Storing only the final failure frame in $\mathcal{B}^-$ is a concise and effective implementation that performs well in our experiments as in Fig.~\ref{fig:main_curves}. This design can also be extended to retain the last few frames of failed episodes when buffer capacity and computation permit. However, including the entire failed trajectory would mistakenly penalize valid intermediate states that also appear in successful data, which would unnecessarily suppress good behaviors.

\paragraph{Observation.}
The policy observation contains two RGB camera
(wrist-view and base-view) and 9-D proprioception
(end-effector position, quaternion, and gripper state). The trainable policy
encoder fuses the two camera views into a 128-D visual embedding, which is
concatenated with proprioception to form the state input for the residual actor
and critic during RL finetuning.

\paragraph{Behavior-cloning pretraining.}
Before online finetuning, we train one behavior-cloning policy per task from
the corresponding offline demonstrations. Each policy is optimized with Adam
(learning rate $10^{-4}$, weight decay $10^{-6}$) and a cosine schedule. We
maintain an exponential moving average of the weights with decay 0.995 and use
the EMA checkpoint as the initial policy for all reward baselines. Thus sparse
reward, RoboMeter, and HiRE start from exactly the same pretrained policy in
each task and seed.

\paragraph{Policy architecture.}
All baselines share the end-to-end trainable policy architecture summarized in
Table~\ref{tab:impl_bc}: a ResNet-18 with spatial softmax for visual encoding
and a 1D UNet action head. The MimicGen tasks use a diffusion policy~\citep{chi2023diffusion_policy}
with $\epsilon$-prediction with 50 DDPM steps;
RoboMimic \texttt{Tool\_Hang} uses a flow-matching policy with velocity prediction and
20-step Euler integration.

\begin{table}[ht]
    \centering\small
    \renewcommand{\arraystretch}{1.15}
    \caption{Policy architecture.}
    \label{tab:impl_bc}
    \begin{tabular}{ll}
    \toprule
    Component & Value \\
    \midrule
    Visual encoder & ResNet-18 + spatial-softmax, trained end-to-end \\
    Keypoints / fusion & 32 keypoints/cam, 64-D/cam $\rightarrow$ fused 128-D \\
    Action head & 1D UNet, base 128, multipliers $[1,2,2]$ \\
    Horizons & $T_a=8$, $T_o=1$, $T_{\mathrm{img}}=1$ \\
    \bottomrule
    \end{tabular}
\end{table}

\subsection{Real-Robot Implementations, Setups, and Results}\label{app:real}
\begin{figure}[ht]
    \centering
    \includegraphics[width=\linewidth]{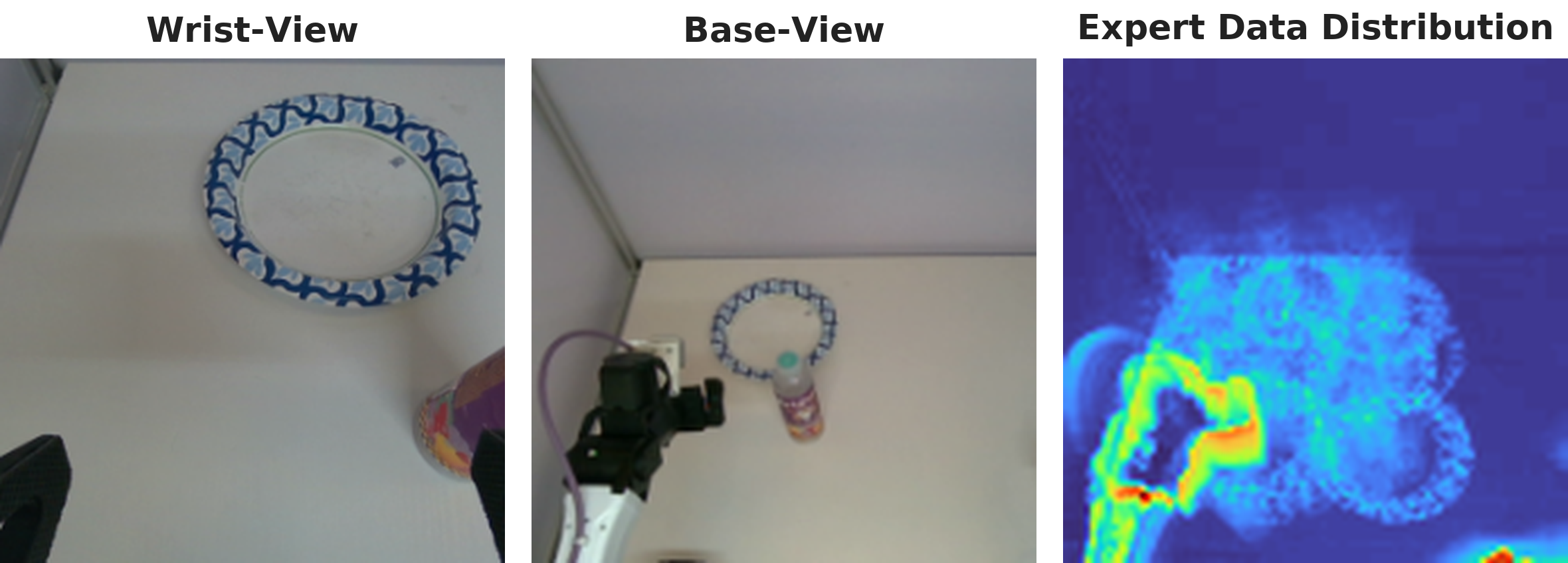}
    \caption{
        \textbf{Wrist view, base view, and visualization of offline expert data.}
    }
    \label{fig:real_robot_setup_training_distribution}
    \vspace{-8pt}
\end{figure}

\textbf{Hardware setup.}
Our real-robot experiments are conducted on the I2RT YAM arm, a 6-DoF manipulator equipped with a parallel linear4310 gripper.
We adopt the teleoperation workstation configuration, which ships with a leader–follower system that allows a human operator to collect expert demonstrations through teleoperation.
Visual observations are captured by two Intel RealSense cameras, with a base-view camera providing a global top view of the workspace (base camera) and a wrist-view camera mounted on the end-effector providing a close-up egocentric view of the gripper and manipulated objects (wrist camera), as shown in Fig.~\ref{fig:real_robot_setup_training_distribution} that use \texttt{Bottle\_Pick-and-Place} as an example. On real hardware, successes and failures are directly judged by human operators.

\begin{figure}[ht]
    \centering
    \includegraphics[width=\linewidth]{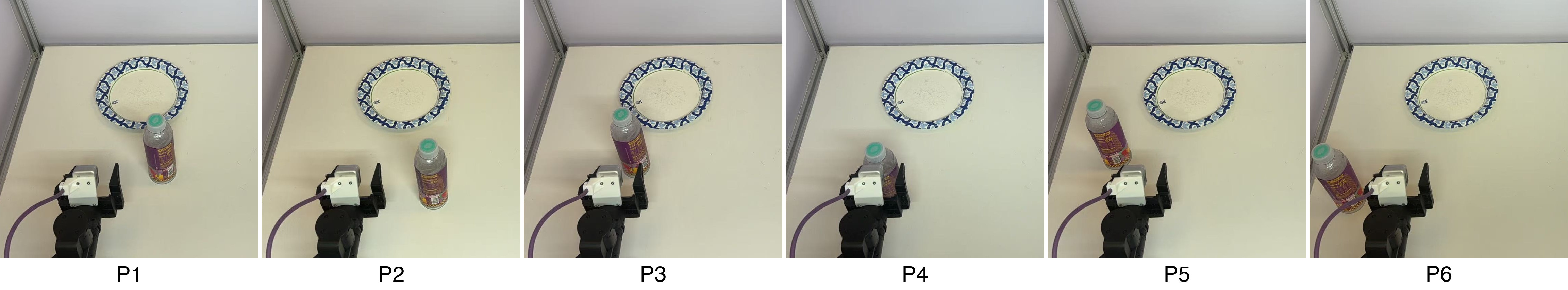}
    \caption{
        \textbf{Six fixed initial bottle positions (P1--P6) used for real-robot evaluation.}
        Each panel shows the base-view camera observation at the start of an episode, with the bottle placed at one of the six predefined table positions. 
    }
    \label{fig:real_robot_init_positions}
    \vspace{-8pt}
\end{figure}

\textbf{Task descriptions.}
\texttt{Bottle\_Pick-and-Place} requires the robot to localize a bottle, grasp it reliably, transport it to a target plate, and release it accurately. 
The complexities of this task are two-fold: 
(1) \textbf{Coarse-to-fine compliance and precision:} The task demands a seamless integration of large-scale macro-movements (e.g., reaching for the bottle or transporting it toward the plate) and high-precision micro-manipulation (e.g., at the moments of grasping and placing). Crucially, because both the plastic bottle and the paper plate are highly deformable, the policy must maintain compliant control during grasping and releasing to prevent the objects from bouncing away or slipping.
(2) \textbf{Spatial generalization:} We evaluate the policies across six different bottle initial positions spanning a wide tabletop workspace (as shown in Fig.~\ref{fig:real_robot_init_positions}), testing the robust generalization of HiRE rewards. We randomize the initial bottle positions within a small range approximately of $\pm10$cm on the table.
We elaborate on the performance of HiRE on each initial position in Tab.~\ref{tab:real_robot_results}.

\texttt{Towers\_of\_Hanoi} requires the robot to move the green disk from the rightmost rod to the middle rod of a Hanoi platform, demanding a sequence of high-precision manipulations. Although the initial setup is standardized across trials for reproducibility, the Hanoi base is not anchored to the tabletop. 
Consequently, execution uncertainties often cause the main platform to shift relative to the workspace, while the removable top cover can independently slide relative to the base. 
These unconstrained spatial shifts introduce severe dynamic perturbations and non-recoverable error states, where even a minor contact alignment error can render the task physically unrecoverable, imposing strict demands on real-time policy precision and feedback robustness.



\textbf{Policy training and evaluation.}
We collect diverse 117 teleoperated expert demonstrations for \texttt{Bottle\_Pick-and-Place} (the distribution density is visualized in Fig.~\ref{fig:real_robot_setup_training_distribution} with colors) and 50 for \texttt{Towers\_of\_Hanoi} to initialize base diffusion policies, following the exact implementation recipe of DICE-RL~\citep{sun2026prior}. For online finetuning, the pre-trained diffusion policy first executes 20 warmup episodes distributed evenly across the designated initial positions. We then conduct policy optimization for 2,000 training steps with a batch size of 256. To maintain a stable update-to-data (UTD) ratio during interaction, the policy samples 10 new episodes after every 1,000 training steps in subsequent training loops, ensuring a consistent training stability and data exploitation.

We evaluate HiRE, sparse-reward and RoboMeter baselines on the YAM robot arms under the same setup.
All methods start from the same pre-trained policy and 20 warmup episodes; the only difference is the reward signal used during finetuning.
For all methods, a human annotator labels each episode as success or failure during execution.
For HiRE, we set positive buffer size to 4096 and negative buffer size to 320. We add all the online successful episodes into the positive buffer.
We add the last 16 frames of online failed episodes into the negative buffer.
We set $\kappa^+=10$, $\kappa^-=1$, $\lambda=0.1$, and batch size 256. All real-robot experiments are conducted on a single NVIDIA RTX 5090 GPU.
In Fig.~\ref{fig:real_curve}, the reported success rates are evaluated across 30/15 trials for each checkpoint on \texttt{Bottle\_Pick-and-Place}/\texttt{Towers\_of\_Hanoi}.

\begin{table}[ht]
\centering\small
\renewcommand{\arraystretch}{1.15}
\caption{\textbf{Detailed Real-robot evaluation rollout success rates (HiRE) on \texttt{Bottle\_Pick-and-Place}.}
Each policy is evaluated at six fixed initial positions (P1--P6) with five trials per
position, for $30$ trials in total. Each cell reports the number of successful attempts out of five, and the last two columns report the overall
success count and success rate. BC denoted the pre-trained base policy; the remaining rows are checkpoints taken at increasing numbers of
online finetuning episodes.}
\label{tab:real_robot_results}
\vspace{10pt}
\begin{tabular}{lcccccccc}
\toprule
Policy & P1 & P2 & P3 & P4 & P5 & P6 & Success & Rate \\
\midrule
BC              & 2 & 0 & 0 & 2 & 0 & 2 & 6/30  & 20.0\% \\
\midrule
Online ep.\ 30  & 1 & 2 & 1 & 0 & 1 & 1 & 6/30  & 20.0\% \\
Online ep.\ 40  & 4 & 4 & 2 & 1 & 1 & 0 & 12/30 & 40.0\% \\
Online ep.\ 50  & 3 & 5 & 2 & 1 & 2 & 3 & 16/30 & 53.3\% \\
Online ep.\ 60  & 3 & 5 & 3 & 3 & 3 & 3 & \textbf{20/30} & \textbf{66.7\%} \\
Online ep.\ 70  & 4 & 3 & 3 & 4 & 3 & 3 & \textbf{20/30} & \textbf{66.7\%} \\
\bottomrule
\end{tabular}
\end{table}

\begin{table}[t]
\centering\small
\caption{DICE-RL finetuning hyperparameters.}
\label{tab:impl_rl}
\begin{tabular}{lp{0.62\linewidth}}
\toprule
Setting & Value \\
\midrule
Residual actor & 3-layer MLP, hidden 1\,024, GELU, LayerNorm; 56-D noise \\
Action decoding & DDIM~\citep{song2020ddim}, 10 steps \\
Critic & ensemble of 10; 16 candidates, \texttt{max\_q\_min} selection \\
Collection & 100 steps/iter $\times$ 8 parallel envs \\
Updates per iter & 20, batch 256 \\
Actor : critic updates & 1 : 2; target soft update $\tau = 0.01$ \\
Optimizer & Adam, lr $10^{-4}$, weight decay $10^{-5}$, cosine warm-up \\
Returns & 5-step, $\gamma = 0.99$ \\
Replay buffer & up to $2{\times}10^{6}$ transitions \\
Evaluation & every 200 steps, 10 episodes $\times$ 10 envs \\
\bottomrule
\end{tabular}
\end{table}

\subsection{Method Implementations}\label{sec:impl_reward_baselines}

 
The main comparison in Fig.~\ref{fig:main_curves} evaluates HiRE against four
reward baselines: sparse reward, RoboMeter, TOPReward, and GCR.
All reward methods are integrated into the same DICE-RL
backbone~\citep{sun2026prior}; only the reward computation
differs across methods. 
DICE-RL freezes the pretrained generative policy
as an action proposal model and learns a residual actor together with an
ensemble critic. We keep all optimization, replay, exploration, and evaluation
settings fixed across reward baselines; only the scalar reward stored for each
transition changes. Table~\ref{tab:impl_rl} lists the hyperparameters.
For visibility, curves with zero success throughout
their recorded evaluations are extended at $0\%$ to the midpoint of the
displayed training range.

\paragraph{HiRE reward.}
HiRE keeps the same sparse success signal as the sparse baseline and
augments it with the control-aware potential $\Phi_{\text{HiRE}}$ of
Eq.~\ref{eq:hire_logsumexp}, edited online from hindsight rollouts. The potential
is built on a frozen DINOv2~\citep{oquab2024dinov} (\texttt{dinov2\_vits14},
ViT-S/14) encoder $\phi$, which we evaluate inside the main training process on
both camera views. Each image is resized to $224{\times}224$, normalized with
ImageNet statistics, and encoded into $P$ normalized patch tokens. The
state--reference similarity $\text{sim}_\phi$ in Eq.~\ref{eq:hire_logsumexp} is
the patch-averaged cosine similarity which is averaged across the two cameras.
\begin{equation}
    \text{sim}_\phi(s,\, g)
        = \frac{1}{P}\sum_{p=1}^{P}
          \frac{\phi_p(s)^\top \phi_p(g)}
               {\|\phi_p(s)\|\,\|\phi_p(g)\|},
\end{equation}

The positive buffer $\mathcal{B}^+$ and negative buffer $\mathcal{B}^-$ of
Eq.~\ref{eq:hire_logsumexp} are instantiated as follows. The positive buffer
$\mathcal{B}^+$ combines an offline component, constructed once from the expert
demonstrations with frame stride 5, and an online FIFO component of capacity 128
that stores terminal frames from online successful rollouts; the two components are mixed
with ratio 0.5 when sampling. The negative buffer $\mathcal{B}^-$ is an online
FIFO buffer of capacity 128, populated only with the terminal frames of failed
rollouts. For online positives, retaining only the successful terminal frame
avoids reinforcing exploratory or suboptimal intermediate behaviors, while
offline expert demonstrations provide positive references along the task
trajectory. For online negatives, retaining only the failed terminal frame
avoids penalizing valid intermediate behaviors that may precede failure.

Finally, $\Phi_{\text{HiRE}}$ is integrated with the sparse reward through the
potential-based reward shaping (PBRS) form of Eq.~\ref{eq:hir_final}:
\begin{equation}
    R_{\text{HiRE}}(s,a,s^\prime)
    = R_{\text{sparse}}(s,a,s^\prime)
    + \frac{w_{\text{dense}}(t)}{T_a}
    \bigl(\gamma\Phi_{\text{HiRE}}(s^\prime)-\Phi_{\text{HiRE}}(s)\bigr).
\end{equation}
We use shaping discount $\gamma=0.99$ and normalize the shaping term by the
action-chunk length $T_a=8$. To keep the dense signal from
interfering with an already competent policy, we down-weight the shaping term by
a sliding-window weight $w_{\text{dense}}(t) = 1 - \mathrm{SR}_{\text{recent}}(t)$,
where $\mathrm{SR}_{\text{recent}}(t)$ is the success rate over the most recent
completed episodes. As the success rate rises, $w_{\text{dense}}$ decays toward
$0$, so HiRE shapes the reward most strongly early in finetuning and gradually
hands control back to the sparse task signal. Following RLPD~\citep{ball2023efficient}, replayed expert
demonstrations keep the sparse environment reward only and are not relabeled with
the online contrastive reward, which keeps the learning signal clean.

\paragraph{Sparse reward.}
The sparse baseline uses the native binary success signal from the benchmark:
\begin{equation}
    R_{\text{sparse}}(s,a,s^\prime)
    =
    \mathds{1}\{s^\prime\ \text{is successful}\}.
\end{equation}
All non-terminal and failed terminal transitions receive $R=0$, and a
successful terminal transition receives $R=1$. 

\paragraph{RoboMeter.}
The RoboMeter~\citep{liang2026robometer} baseline uses the publicly released \texttt{RoboMeter-4B}
checkpoint as a frozen vision--language dense reward
model. Following the official LIBERO-style inference wrapper, we run RoboMeter
as a separate GPU service and query it from the RL trainer through batched HTTP
requests. This decouples VLM inference from policy optimization and keeps the
reward model fixed throughout finetuning. For each parallel environment, we
maintain an online RGB frame buffer from base-view camera for
reward inference.
The episode prefix is
evenly subsampled along time to at most 16 frames, always retaining the first
and most recent frames. The resulting clip is conditioned on the corresponding
natural-language task instruction.

The reward signal is obtained from RoboMeter's progress head. Given the progress
curve $\hat{p}(s^\prime)$ predicted from the episode prefix ending at the next
state $s^\prime$, we take its final prediction and clip it to $[0,1]$:
\begin{equation}
    p(s^\prime) = \mathrm{clip}\bigl(\hat{p}(s^\prime)[-1],\,0,\,1\bigr),
\end{equation}
Throughout this paper, RoboMeter denotes the relative-progress reward, which
adds the change in predicted progress to the sparse success signal:
\begin{equation}
    R_{\text{RM}}(s,a,s^\prime)
    =
    R_{\text{sparse}}(s,a,s^\prime)
    +
    \bigl(p(s^\prime)-p(s)\bigr).
\end{equation}
Here $p(s)$ and $p(s^\prime)$ denote the cached progress estimates before and
after the transition, respectively. We reset the progress estimate to zero at
the initial state of every episode. We query RoboMeter every $N=4$ action chunks
and cache the most recent progress estimate between queries. This reward
therefore measures progress increments rather than repeatedly rewarding the
same absolute progress level.

\paragraph{TOPReward~\citep{chen2026topreward}.}
The TOPReward baseline uses the frozen
\texttt{Qwen3-VL-8B-Instruct} model and follows the official scoring
implementation. At each action chunk, we uniformly sample at most 16 base-view
frames from the current episode prefix, preserving their temporal order, and
condition the resulting clip on the natural-language task instruction.

The model scores the log-probability of an affirmative task-completion
judgment without first generating a trajectory description. Let $\ell(s)$
denote the raw score for the episode prefix ending at state $s$, computed with
mean token reduction. We
convert successive scores into a clipped difference reward:
\begin{equation}
    R_{\text{TOPReward}}(s,a,s^\prime)
    = \mathrm{clip}\bigl(2(\ell(s^\prime)-\ell(s)),\,-1,\,1\bigr).
\end{equation}
The first query of each episode receives zero reward. This signal replaces the
sparse environment reward for online rollouts. We apply the difference directly
to raw scores, without min--max normalization across trajectory prefixes.

\paragraph{GCR~\citep{biza2025robot}.}
We implement the simple-contrastive variant \texttt{GCR(SC)}~\citep{biza2025robot}
following the method and implementation details in the original paper and its
appendix, as no official implementation was publicly available to us.
The reward model is initialized from the pretrained LIV~\citep{ma2023liv}
ResNet-50 checkpoint, with the full visual backbone finetuned during reward
learning. We use RGB observations from the base-view camera.

The training objective combines the VIP temporal loss with simple goal
contrastive learning: it increases similarity between successful goals from
different demonstrations and decreases similarity between successful goals
and online failure states. Successful goals and negative examples are sampled
from the terminal segments of offline successful demonstrations and online
failed episodes, respectively. The model is first finetuned on offline
demonstrations and then updated online as the policy collects new experience.

For each episode, we sample a successful goal frame $g$ and keep it fixed.
Let $c_\theta(s,g)$ denote cosine similarity in the learned representation
space, and define $\Phi_{\text{GCR}}(s;g)=(c_\theta(s,g)+1)/2$. The online
reward follows the difference form and reward scaling described in the paper:
\begin{equation}
    R_{\text{GCR}}(s,a,s^\prime)
    = 10\bigl(R_{\text{sparse}}(s,a,s^\prime)
    + \Phi_{\text{GCR}}(s^\prime;g)-\Phi_{\text{GCR}}(s;g)\bigr).
\end{equation}
We score the current and next observations with the same predictor snapshot.
Replayed expert demonstrations retain only the sparse environment reward,
scaled by the same factor of 10.


\section{Additional Mathematical Interpretations}
Mathematically, the precise landscape topography across these situations is regulated by the concentration parameter $\kappa$. The parameter $\kappa$ functions exactly as the inverse temperature in the standard LogSumExp operator~\citep{boyd2004convex}, which controls the smoothness of the maximum operator over the support sets. Under the hard-assignment limit where the representation distributions become highly concentrated ($\kappa \to \infty$), the smooth expectation is dominated entirely by the single nearest neighbor in each buffer, simplifying HiRE into a clear geometric subtraction: $\lim_{\kappa \to \infty} \Phi_{\text{HiRE}}(s) = \max_{g^+ \in \mathcal{B}^+} \text{sim}_\phi(s, g^+) - \lambda \max_{g^- \in \mathcal{B}^-} \text{sim}_\phi(s, g^-)$.
While this hard limit provides an intuitive visualization of the underlying boundary subtraction, our final implementation utilizes the smooth version ($\kappa < \infty$). This choice delivers denser, continuous gradients and leverages the full empirical data distribution within the interaction history to stabilize policy finetuning.

\section{Additional Results}\label{app:addition}

\subsection{Ablation Results}
We organize the ablations into four groups:
(i)~\textbf{foundation representation}, swapping the encoder to evaluate HiRE
with different foundation representations; (ii)~\textbf{hyperparameter sensitivity}, varying the
contrastive weight $\lambda$ and negative-buffer capacity to examine HiRE's
insensitivity to changes in these settings;
(iii)~\textbf{reward composition}, comparing full HiRE with direct addition of
the edited potential (without PBRS) and direct addition of raw goal similarity
(without hindsight editing or PBRS);
(iv)~\textbf{buffer design}, disabling the negative buffer to measure how much
the failure data contributes to reward correction on top of the success data.


\begin{figure*}[ht]
\centering
\includegraphics[width=\linewidth]{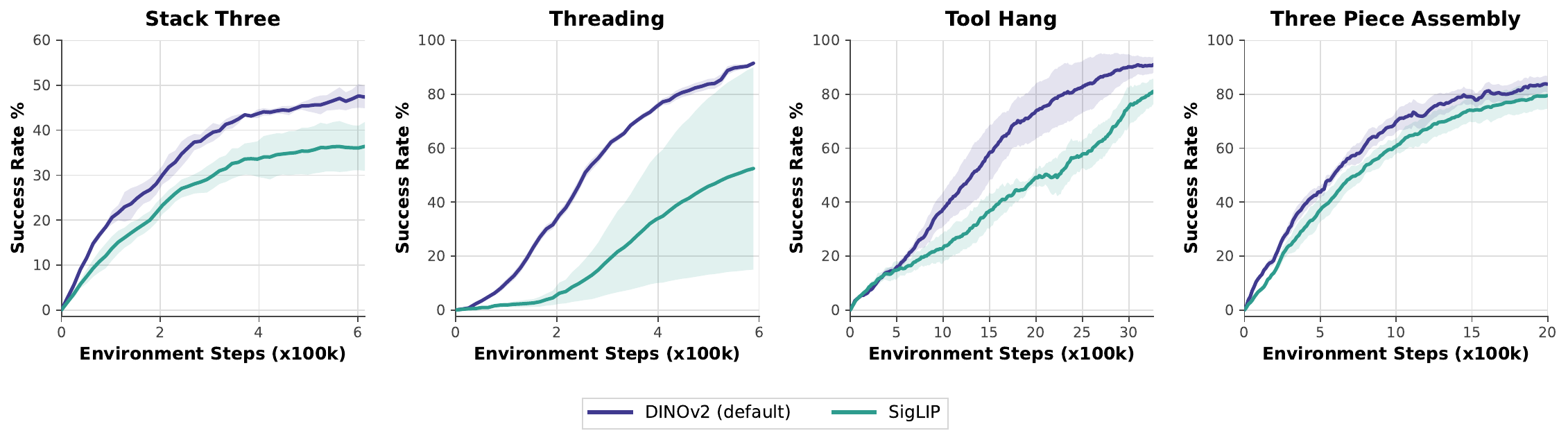}
    \caption{
        \textbf{Foundation representation ablations.}
        We compare HiRE with frozen DINOv2 and SigLIP encoders.
        Curves show mean success rates, with shaded regions denoting variability
        across available random seeds.
    }
\label{fig:ablation_encoder_all}
\vspace{-10pt}
\end{figure*}

\paragraph{Foundation Representation Model.}
HiRE only requires a frozen encoder $\phi$ that maps visual observations to
patch-level features, so it can accommodate different foundation
representations. We replace the default DINOv2 encoder with the SigLIP visual
encoder~\citep{tschannen2025siglip} (ViT-B/16), retaining patch-averaged cosine
similarity and hindsight contrastive editing.
As shown in Fig.~\ref{fig:ablation_encoder_all}, \textbf{both encoders provide
useful dense guidance across all four tasks}, while DINOv2 consistently achieves
higher success rates. The difference
is task-dependent: SigLIP approaches DINOv2 on \texttt{Tool\_Hang} and
\texttt{Three\_Piece\_Assembly}, but exhibits slower improvement on
\texttt{Stack\_Three} and particularly \texttt{Threading}, where its variability
across seeds is also larger. These results suggest that HiRE is versatile across
foundation representations, while the choice of encoder still affects sample
efficiency and the reliability of the resulting reward landscape.

\begin{figure*}[ht]
\centering
\includegraphics[width=\linewidth]{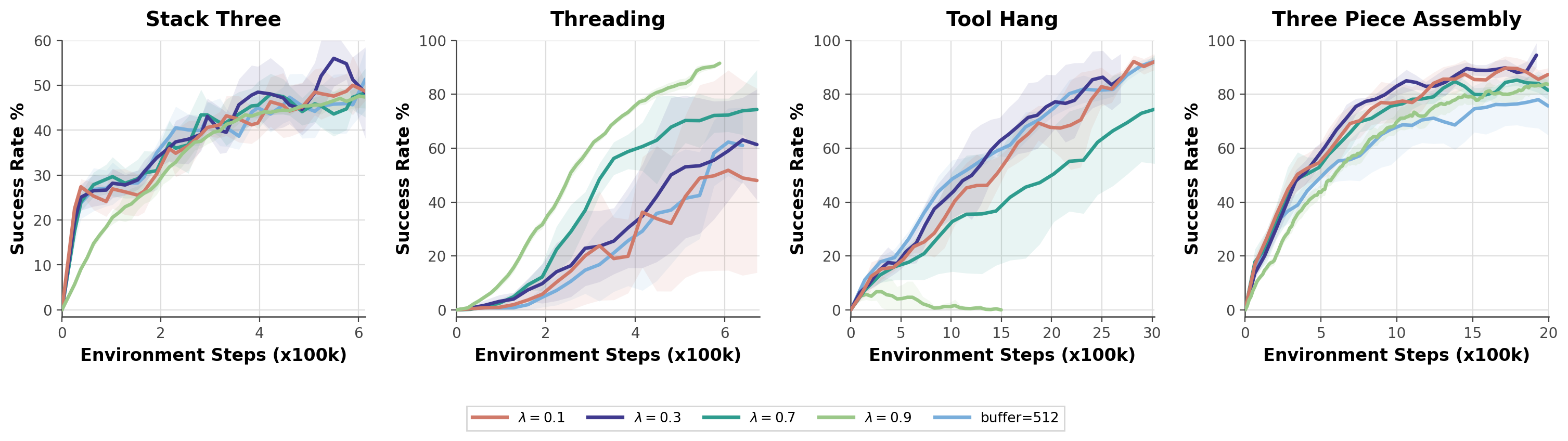}
    \caption{
        \textbf{Hyperparameter sensitivity across four simulation tasks.}
        We compare contrastive weights $\lambda\in\{0.1,0.3,0.7,0.9\}$ and an
        additional configuration with negative-buffer capacity 512
        (\texttt{buffer=512}). Curves are averaged over three random
        seeds, with shaded regions denoting variability across seeds.
    }
\label{fig:ablation_hyparam}
\vspace{-10pt}
\end{figure*}

\paragraph{Hyperparameter Sensitivity.}
We evaluate sensitivity by varying the contrastive weight
$\lambda\in\{0.1,0.3,0.7,0.9\}$ and testing a negative-buffer capacity of 512.
As shown in Fig.~\ref{fig:ablation_hyparam}, \textbf{HiRE retains effective
learning across a broad range of settings} on \texttt{Stack\_Three} and
\texttt{Three\_Piece\_Assembly}. The variants perform similarly on
\texttt{Stack\_Three}, while all configurations achieve substantial performance improvment on
\texttt{Three\_Piece\_Assembly}, only with some differences in learning speed and
asymptotic performance. The high-precision \texttt{Threading} task shows greater
sensitivity: reducing $\lambda$ or using the larger buffer slows improvement
relative to the default setting. A larger buffer can retain outdated failures
and weaken the focus on current traps, while a smaller $\lambda$ may provide
insufficient repulsion when the failure distribution is broad.
The \texttt{Tool\_Hang} sweep reveals a complementary regime:
$\lambda=0.1$ and $0.3$ provide stronger guidance, whereas $\lambda=0.9$ fails to
sustain improvement. 
Here, it echoes what we reveal in Fig.~\ref{fig:buffer_samples_scatter_plot}: the positive distribution remains highly concentrated, whereas the negative one is much wider and scattered, probably because the failure cases vary, but the successful paradigm is very unique. This empirical finding confirms assigning an even larger concentration factor $\kappa^+$ than to $\kappa^-$, which we discuss in Sec.~\ref{hire}.
Overall, the results support broad parameter
insensitivity on the multi-stage tasks, while highlighting the need for
appropriate failure-based correction on precision-critical tasks.

\begin{figure*}[ht]
    \centering
    \includegraphics[width=\linewidth]{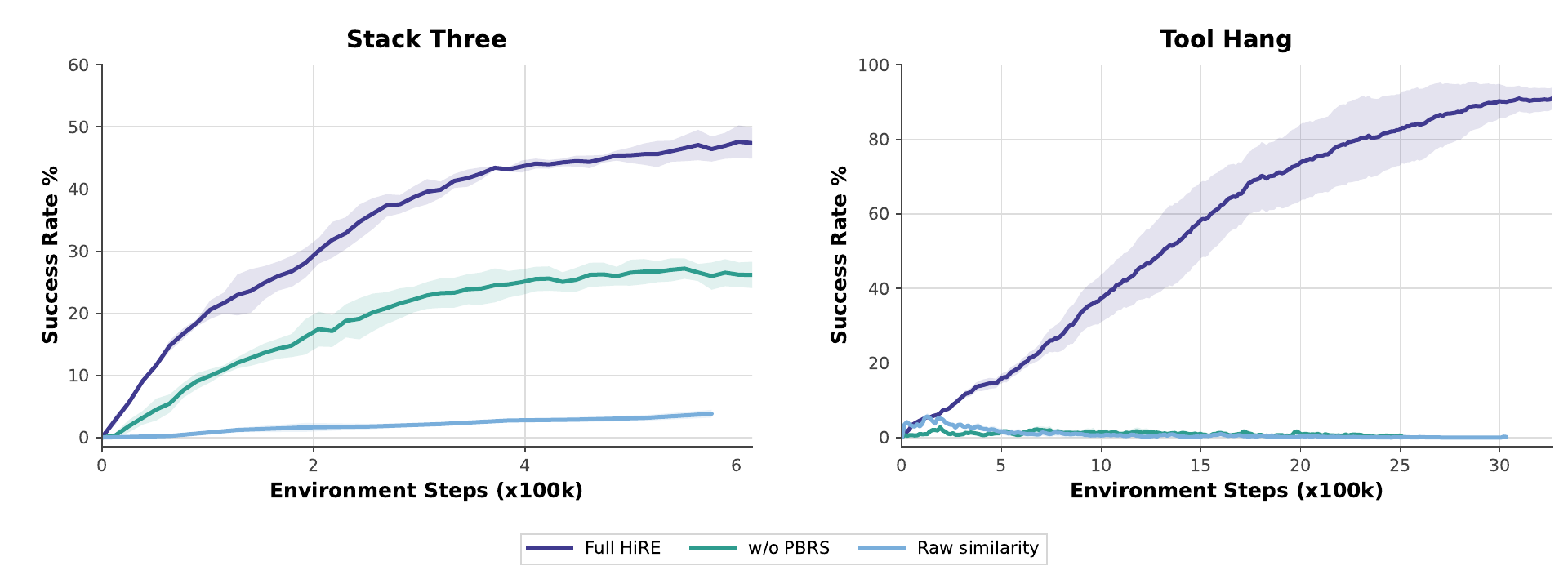}
    \caption{
        \textbf{Reward-composition comparisons on \texttt{Stack\_Three} and
        \texttt{Tool\_Hang}.}
        We compare full HiRE, direct addition of the HiRE potential
        (w/o PBRS), and direct addition of raw goal similarity
        (Raw similarity). Curves are averaged over three random
        seeds, with shaded regions denoting variability across seeds.
    }
    \label{fig:ablation_stack_three_ab}
    \vspace{-10pt}
\end{figure*}

\paragraph{Reward Composition.}
The full HiRE reward combines hindsight contrastive editing
(Eq.~\ref{eq:hire_logsumexp}) with potential-based reward shaping
(Eq.~\ref{eq:hir_final}). Fig.~\ref{fig:ablation_stack_three_ab} compares this
composition with two alternative reward forms:
\begin{itemize}[leftmargin=*]
    \item \textbf{HiRE without PBRS.}
    We add the edited potential directly to the sparse success reward:
    \begin{equation}
        R(s,a,s^\prime) = R_{\text{sparse}}(s,a,s^\prime)
            + w\cdot\Phi_{\text{HiRE}}(s^\prime).
    \end{equation}
    This retains hindsight contrastive editing but removes the
    temporal-difference structure of PBRS.

    \item \textbf{Raw similarity.}
    We replace the edited potential with the raw goal-similarity potential
    $\Phi_{\text{raw}}(s)=\text{sim}_\phi(s,g^+_\star)$, where $g^+_\star$ is the
    final successful frame of an expert demonstration, and add it directly
    to the sparse reward:
    \begin{equation}
        R(s,a,s^\prime) = R_{\text{sparse}}(s,a,s^\prime)
            + w\cdot\Phi_{\text{raw}}(s^\prime).
    \end{equation}
    This uses a fixed visual goal without hindsight correction from successful
    and failed rollouts or the temporal-difference structure of PBRS.
\end{itemize}
Here $w$ denotes the dense-reward coefficient; the equations summarize the
reward forms. The compared runs also differ in reward scaling and replay
settings, so these results compare the resulting reward configurations rather
than isolating a single implementation change.

On \texttt{Stack\_Three}, \textbf{full HiRE achieves both faster improvement
and higher success rates} than the two alternatives. Removing PBRS leads to
slower improvement and an earlier plateau, while raw similarity provides
little learning progress.
The advantage of the variant without PBRS over raw similarity suggests that
hindsight editing provides useful guidance even with direct reward addition.
The further gain from full HiRE is consistent with the benefit of integrating
the edited potential through temporal differences that cancels intermediate terms during intrinsic reward accumulation.

The distinction is sharper on \texttt{Tool\_Hang}, where both alternative
configurations remain near zero while full HiRE continues to improve.
These results suggest that \textbf{combining hindsight editing with PBRS is
especially valuable for contact-rich, high-precision manipulation}: direct
addition of either the edited potential or raw goal similarity does not
reproduce the learning progress achieved by full HiRE.

\begin{figure*}[ht]
    \centering
    \includegraphics[width=\linewidth]{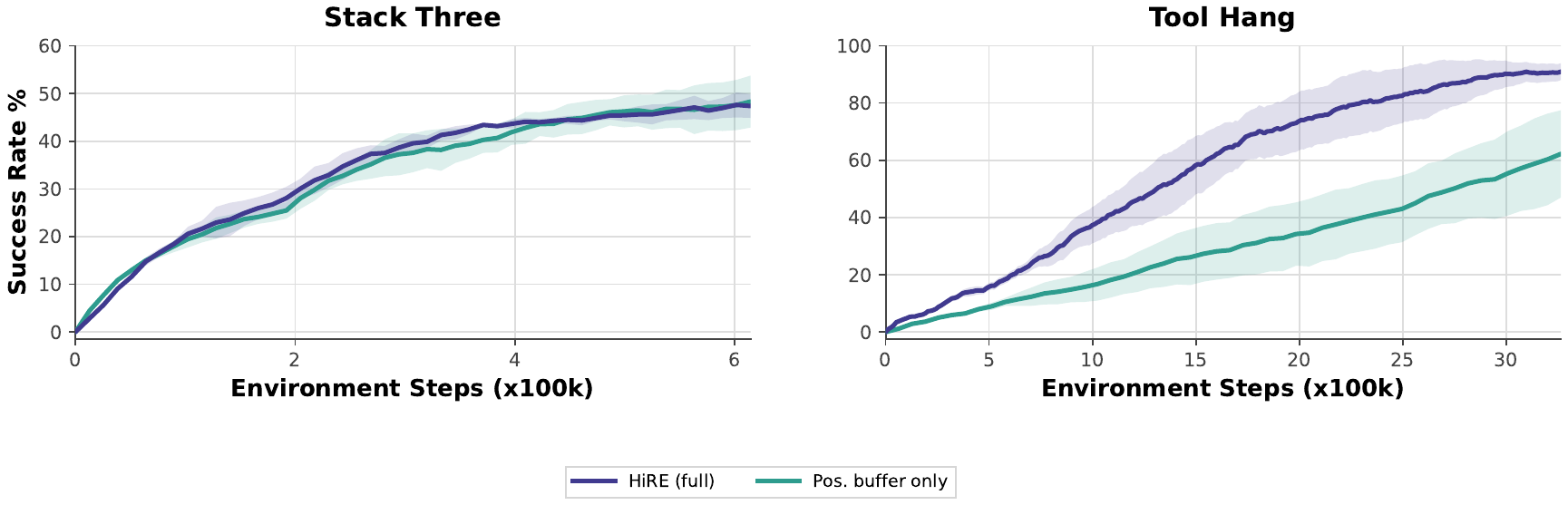}
    \caption{
        \textbf{Buffer-design ablations on \texttt{Stack\_Three} and
        \texttt{Tool\_Hang}.}
        We compare full HiRE with the positive-buffer-only variant, which removes
        the negative buffer and keeps only success-manifold attraction. Curves are averaged over three random
        seeds, with shaded regions denoting variability across seeds.
    }
    \label{fig:ablation_buffer_results}
    \vspace{-8pt}
\end{figure*}

\paragraph{Buffer Design.}
Section~\ref{hire} interprets HiRE as a contrast between two empirical
manifolds: $\mathcal{B}^+$ provides success-dominated attraction that rescues
false negatives, whereas $\mathcal{B}^-$ provides failure-dominated repulsion
against false-positive trap states. To isolate the role of the latter, we
evaluate a \textbf{positive-buffer-only} variant that removes the negative buffer
and keeps only attraction toward the success manifold:
\begin{equation}
    \Phi(s)
    =
    \mathbb{L}^{\kappa^+}_{\mathcal{B}^+}
    \left(\text{sim}_\phi(s, g^+)\right).
\end{equation}
This ablation tests whether the success manifold alone is sufficient, and how
much additional benefit comes from explicit failure awareness.
As shown in Fig.~\ref{fig:ablation_buffer_results}, the positive-buffer-only
variant closely tracks full HiRE on \texttt{Stack\_Three}, and finishes
slightly higher with overlapping variability bands. This suggests that
success-manifold attraction supplies much of the dense guidance on this task.
The benefit of the full contrastive reward is clearer on \texttt{Tool\_Hang}. This task is
contact-rich, high-precision, and multi-stage, so it produces many near-goal
failures that appear visually successful. Such states are exactly the
false-positive traps that attraction to $\mathcal{B}^+$ cannot distinguish, but
that repulsion from $\mathcal{B}^-$ can suppress. Full HiRE improves faster
and reaches above $90\%$, while the positive-only variant improves more slowly
and achieves a lower final success rate. This difference across
the two tasks suggests that the value of the negative buffer grows with task
difficulty and the prevalence of visually deceptive failures.
Overall, the ablation is
consistent with the interpretation in Section~\ref{hire}: the positive buffer
supplies attraction toward successful states, while the negative buffer provides
the complementary correction needed to suppress visually deceptive failures.

\end{document}